\documentclass[a4paper]{article}

\usepackage[latin1]{inputenc}
\usepackage{amsmath}

\usepackage{amsfonts}
\usepackage{color}
\usepackage{stmaryrd}
\usepackage{hyperref}
\usepackage[bottom]{footmisc}
\usepackage{lmodern}
\usepackage{float}
\usepackage{caption}
\usepackage[textsize=small,colorinlistoftodos]{todonotes}

\usepackage[a4paper]{geometry}
\definecolor{rouge}{rgb}{1,0,0}

\newcommand{\defl}{\textbf{\texttt{deflation }}}
\newcommand{\same}{\textbf{\texttt{block\_same\_mu }}}
\newcommand{\diff}{\textbf{\texttt{block\_diff\_mu }}}

\newcommand{\V}{U}
\newcommand{\vv}{u}
\newcommand{\W}{V}
\newcommand{\ww}{v}
\newcommand{\ve}{\nu}
\newcommand{\p}{^{\prime}}
\newcommand{\ns}{n}
\newcommand{\nvg}{g}
\newcommand{\nvs}{p}
\newcommand{\dv}{p}
\newcommand{\nl}{m}
\newcommand{\rank}{\mathrm{rank}}
\newcommand{\diag}{\mathrm{diag}}
\newcommand{\var}{\mathrm{var}}
\newcommand{\pev}{\mathrm{pev}}
\newcommand{\varpca}{\var_{PCA}(\nl)}
\newcommand{\varmu}{\var_{\mu}}
\newcommand{\varmugamma}{\var_{\mu}^{\gamma}}
\newcommand{\vargamma}{\var^{\gamma}}

\newcommand{\varproj}{\mathrm{var}_{proj}}

\newcommand{\varprojqr}{\mathrm{var}_{proj}^{QR}}
\newcommand{\varprojup}{\mathrm{var}_{proj}^{UP}}
\newcommand{\varprojopt}{\mathrm{var}_{proj}^{opt}}

\newcommand{\varsubspace}{\mathrm{var}_{subsp}\,}
\newcommand{\varnorm}{\mathrm{var}_{norm}}
\newcommand{\varnormqr}{\mathrm{var}^{QR}_{norm}}
\newcommand{\varnormup}{\mathrm{var}^{UP}_{norm}}
\newcommand{\polar}{\mathrm{polar}}
\newcommand{\spann}{\mathrm{span}}
\newcommand{\pr}{\mathrm{P}}
\newcommand{\tr}{\mathrm{tr}}

\newcommand{\snvs}{\mathcal{S}^{\nvs}}

\newcommand{\bpnvs}{(\mathcal{B}^{\nvs})^{\nl}}

\newcommand{\stns}{\mathcal{S}_{\nl}^{\ns}}

\newcommand{\R}{I\!\!R}

\newcommand{\E}{ \mathcal{E}}
\newcommand{\EX}{\mathcal{E}^{X}}
\newcommand{\PX}{\mathcal{P}^{X}}

\newcommand{\Fmugamma}{F}

\newcommand{\cqfd}{\mbox{}\hfill\rule{.8em}{1.8ex}}

\newcommand{\egaldef}{\stackrel{\mathrm{def}}{=}}
\newcommand{\egalquest}{\stackrel{\mathrm{?}}{=}}

\newcommand{\rv}{\mathrm{R_{V}}}

\newtheorem{Theorem}{Theorem}[section]
\newtheorem{Definition}[Theorem]{Definition}
\newtheorem{Proposition}[Theorem]{Proposition}
\newtheorem{Lemma}[Theorem]{Lemma}

\author{
  Marie Chavent 
  \thanks{IMB, Universit\'e de Bordeaux, 33400 Talence, France, 
  \newline \hspace*{1,5em} e-mail~:   {\tt marie.chavent@u-bordeaux.fr} (corresponding author)}
  \hspace{0,1em}
  \thanks{Inria Bordeaux Sud-Ouest, 33405 Talence, France}\hspace{0,2em}
  \and Guy Chavent
  \thanks{Inria-Paris, 2 rue Simone Iff, 75589 Paris, France,
   \newline \hspace*{1,5em}  e-mail~:   {\tt guy.chavent@inria.fr}}\hspace{0,2em}
}

\title{Optimal Projected Variance Group-Sparse Block PCA}

\date{\today}

\begin{document}

\begin{titlepage}

\maketitle
\thispagestyle{empty}
\setcounter{page}{0}

\begin{abstract}
We address the problem of defining a group sparse formulation for Principal Components Analysis (PCA) - or its equivalent formulations as Low Rank approximation or Dictionary Learning problems - which achieves a compromise between maximizing the variance explained by  the components and promoting sparsity of the loadings.
So we propose first a new definition of the variance explained by  non necessarily orthogonal components, which is optimal in some aspect and compatible with the principal components situation. Then we use a specific regularization of this variance by the group-$\ell_{1}$ norm to define  a Group Sparse Maximum Variance (GSMV) formulation of PCA.
The GSMV formulation achieves our objective by construction, and has the nice property that the inner non smooth optimization problem can be solved analytically, thus reducing GSMV to the maximization of a smooth and convex function under unit norm and orthogonality constraints, which generalizes  \cite{JNRS2010} to group sparsity.
Numerical comparison with deflation on synthetic data shows that GSMV produces steadily slightly better and more robust results for the retrieval of hidden sparse structures, and is about three times faster on these examples. Application to real data shows the interest of group sparsity for variables selection in PCA of mixed data  (categorical/numerical) .
\end{abstract}


\end{titlepage}

\section*{Introduction}
\label{intro}

Principal Components Analysis (PCA), Low Rank Approximation (LRA), Self-contained Regression-type formulation (REGR) and Dictionary Learning (DL)  are equivalent problems whose solution is the matrix  $Z$ made of the  singular vectors associated to the $\nl \leq \rank A$ largest singular values of a $\ns \times \nvs$ data matrix $A$ which collects $\ns$ samples made each of $\nvs$ entries~:

\begin{equation} 
\label{i10}
\begin{array}{ccl}
  \text{(PCA)}    &  \displaystyle \max_{Z^{T}Z = I_{\nl}} \|AZ\|^{2}_{F} &\\
     \text{(LRA)}   &  \displaystyle \min_{Z^{T}Z =  I_{\nl}}  \|A - AZZ^{T}\|_{F}^{2} & \\
    \text{(REGR)}    & \displaystyle   \min_{B,Z^TZ=I_{\nl}} \| A-ABZ^{T}\|^{2}_{F} &
   \quad \text{ (and \quad $\displaystyle 
    \arg \min_{B} = Z $)} \\
    \text{(DL)}    & \displaystyle   \min_{Y,Z} \| A-YZ^{T}\|^{2}_{F}  &
    \quad \text{ (and \quad  $\displaystyle 
    \arg \min_{Y} =A Z (Z^TZ)^{-1} $)}
\end{array}
\end{equation}
where the subscript $F$ denotes the Frobenius norm. In the PCA context, $Z$ represents the \emph{loadings} and  $Y=AZ$ the components, whereas in the DL context,  $Z$ is the dictionary, and $Y$ the coefficient matrix.

The equivalence of PCA, LRA and REGR follows from the identity~:
\begin{equation}
\label{i12}
 \|AZ\|_{F}^{2}  + \|A - AZZ^{T}\|_{F}^{2} = \|A\|_{F}^{2} \quad \text{as soon as} \quad Z^{T}Z = I_{\nl} 
\end{equation}
and that of PCA and DL follows from~:
\begin{equation}
\label{i14}
R(A,Z)   +  \min_{Y} \| A-YZ^{T}\|^{2}_{F} = \|A\|^{2}_{F} \ ,
\end{equation}
where $R(A,Z)$ is the Generalized Rayleigh Quotient~:
\begin{equation}
\label{i16}
R(A,Z) = \tr\{ Z^{T}A^{T}AZ(Z^{T}Z)^{-1}\} \ .
\end{equation}
But due to noise the eigenvectors vectors of the sample covariance matrix $A^{T}A$ can be very different from those of the underlying population covariance matrix, especially for  high dimensional problems where 
$\nvs \gg \ns$, and the need of adding information arises. This is usually done by modifying one of the formulations \eqref{i10} in order to promote sparsity, limit total variation and sometimes promote smoothness of the loadings/dictionnary vectors $Z$.

Along with the formulations derived from  \eqref{i10}, probabilistic approaches have been developed with the objective of producing asymptotically optimal estimators, e.g. 
\cite{cai2013} and  \cite{vu2013minimax}, who use joint sparsity constraints between the loadings for the determination of the principal eigensubspace, and provide convergence rates,  
\cite{deshpande2016sparse} who use covariance thresholding,
and the review paper by \cite{cai2016estimating}.

Among the works based on the PCA formulation, we can cite 
\cite{jolliffe2003modified}, who add an exterior penalization to promote sparsity, \cite{JNRS2010}, who introduce the G-power method to implement $\ell_{0}$ or $\ell_{1}$ regularization, \cite{richtarik2012alternating}, who search for sparse leading components by introducing an auxiliary variable to compute $\|Az\|$, and an alternate maximization scheme for the  $\ell_{0}$ or $\ell_{1}$ penalized version,
\cite{ma2013} uses orthogonal iteration with thresholding  to compute the principal subspace, 
\cite{yuan2013truncated} determine  sparse eigenvectors under $\ell_{0}$ constraint
using a truncated power method and deflation,
and \cite{croux2013ro}  combine the maximization of a robust variance penalized with an $\ell_{1}$ term with deflation and a ``grid algorithm'' for the optimization on the unit sphere.

Examples of approaches based on the LRA formulation can be found in 
\cite{SH2008}, who use a rank one approximation with $\ell_{1}$ regularization to compute sparse loadings one at a time by deflation, and in 
\cite{wang2017regularized} where the LRA formulation is used to determine sparse and
smooth loadings in the context of spatial data.

The REGR formulation allows to implement sparsity on the unconstrained variable $B$, which is an advantage over the three previous formulations. It was introduced by \cite{zou2006sparse}, who solve it by  alternate minimization with respect to $B$ (non smooth optimization)  and $Z$ (ascent gradient on a variety), 
and used by \cite{khan2015joint}, who proposed a joint group sparse approach for hyperspectral imaging, where the same sparsity pattern is required on all loadings.

Finally, the dictionary learning formulation DL, which allows to implement the sparsity constraints directly on unconstrained loadings $Z$, has experienced a fast growing due to the development of image analysis. It was introduced, in a slightly different form, by \cite{zhang2002low}  in the context of sparse low rank approximation, who solved it by an heuristic thresholded SVD combined with deflation. 
Dictionary learning with structured sparsity is considered by
\cite{pmlr-v9-jenatton10a} and \cite{bach2012structured}, who examine various adapted convex and non convex penalization terms. 
\cite{guo2015spatially} use a spatially weighed DL formulation with a 
$\ell_{1}$ penalization for image classification, and
 \cite{de2017structured} consider penalization by both structured $\ell_{1}$ norm and total variation for image analysis.
 
 The above mentioned works represent a small part of the relevant literature, but they are cited here to illustrate the fact that all method derived from the formulations \eqref{i10} achieve a compromise between sparsifying $Z$ and maximizing $\|AZ\|^{2}$ (c.f. \eqref{i12} for PCA, LRA and REGR formulations) or $R(A,Z)$ (see \eqref{i14} for the DL formulation). 
 
But sparse loadings and the corresponding components are not orthogonal, and none of these quantities is anymore a satisfactory measure of the variance explained by the components, as shown in Section \ref{adjusted optimal and orthogonal variances} for 
 $\|AZ\|^{2}$ and Lemma \ref{lem 1} for $R(A,Z)$. So we take in this paper a slightly different point of view~: we first define a variance measure for non necessarily orthogonal components, and then use it as a starting point  for the definition of sparse PCA formulations, thus ensuring directly a compromise between sparsity and explained variance.

To this effect, we study first the problem of defining the variance explained by a set of non-necessarily orthogonal components.
Two definitions have been proposed in the literature~: the {\em adjusted variance}  of  \cite{zou2006sparse} and the {\em total variance} of \cite{SH2008}, 
which we complement by two definitions based on projection ({\em polar variance} and {\em optimal variance}) and two definitions based on normalization.
We introduce then a set of properties to be satisfied
by any variance definition in order to ensure compatibility with the case of PCA, where the components are orthogonal, and prove that five out of the six above definitions satisfy these compatibility conditions. 
Numerical experimentation confirms the theoretical results, and shows that the ranking of components by variance is essentially independant of the chosen  definition.
As a conclusion of this study, we propose to {\em define the variance of any set of components} $Y=[y_{1}\dots y_{\nl}]$
by the {\em optimal projected variance}~:
\begin{equation}
\label{i1}
\var Y =  \max_{X^{T}X=I_{\nl}} \sum_{j=1\dots\nl} \langle y_{j}\,,x_{j}\rangle^{2} \ ,
\end{equation}

We can now add to the four equivalent formulations  \eqref{i10} a fifth equivalent \emph{Maximum Variance} formulation~:
\begin{equation}
\label{i17}
  \text{(MV)}  \quad \quad \displaystyle \max_{ \|z_{j}\|=1 , j=1\dots \nl } \var(AZ) \ .
\end{equation}

The next step is to use \eqref{i17} to define a sparse PCA formulation. We have chosen to implement group sparsity, where the groups form a partition of $\{1 \dots \nvs \}$, for two reasons~: first, it is useful for the
analysis of mixed data, which contain both numerical and categorical variables 
[\cite{hill1976}, \cite{de1980}, \cite{kiers1991}, \cite{pages2004}, \cite{pages2014}, \cite{chavent2012orthogonal}], and
require an algorithm able to set simultaneously to zero all loading coefficients associated to groups of binary variables used to represent the levels of a categorical variable. 
Second, this choice simplifies greatly, when properly implemented, the numerical resolution of the maximization problem, avoiding the need of solving non smooth optimization problems.

So in order to derive from  \eqref{i17} a  \emph{group-sparse} PCA formulation, 
we choose a group-$\ell_{1}$ norm $\|z_{j}\|_{1}$ and sparsity regularization parameters $\gamma_{j}$, but, instead of substracting  $\sum_{j=1\dots\nl} \gamma_{j}\|z_{j}\|_{1}$ from the MV formulation \eqref{i17} as it is usually done, we define a \emph{perturbed variance} 
$\vargamma(AZ)$ by~:
\begin{equation}
\label{i5}
  \vargamma (AZ) = \max_{X^{T}X=I_{\nl}} \sum_{j=1\dots\nl}
   \big[ \langle Az_{j},x_{j}\rangle -\gamma_{j}\|z_{j}\|_{1} \big]_{+}^{2}  \ ,
\end{equation}
and define the \emph{Group Sparse Maximum Variance} formulation (compare 
with \eqref{i17}) by~:
\begin{equation}
\label{i20}
\hspace{-0,45em} \text{(GSMV)}\quad 
\displaystyle \max_{ \|z_{j}\|=1 , j=1\dots \nl } \vargamma(AZ) 
=  \max_{\|z_{j}\|=1 , j=1\dots \nl}  \max_{X^{T}X=I_{\nl}} \sum_{j=1\dots\nl}
   \big[ \langle Az_{j},x_{j}\rangle -\gamma_{j}\|z_{j}\|_{1} \big]_{+}^{2} \ .
\end{equation}
By construction, this formulation achieves the desired balance
between maximizing the variance of the (non-necessarily orthogonal) components and  promoting group-sparsity of the (non-necessarily orthogonal ) unit norm loadings.
And it has the nice  feature  that the (non smooth) optimization with respect to $Z$ can be solved analytically, thus reducing resolution of GSMV to
 the maximization of a smooth convex function over all $X$ such that $X^{T}X=I_{\nl}$.
 In the case of scalar variables, with $\nvs$ groups made of one variable each, this maximization problem coincides with the remarkable \emph{block $\ell_{1}$ formulation} of \cite{JNRS2010}, from which we borrow the maximization gradient algorithm (page 526).
 
We propose a strategy for the initialization of the algorithm and for the choice of the  parameters $\gamma_{j}$, aimed at balancing the sparsifying effort equally on all loadings.

The performance of the GSMV algorithm is compared to deflation on synthetic data with known underlying group sparsity structure. Then we illustrate the importance of the group-sparsity capability of the algorithm on a real data set containing numerical variables and categorical variables (mixed data).

The paper is organized as follows: Section~\ref{pca} recalls the notations for PCA (Principal Component Analysis),
Section~\ref{how to define the explained variance} is devoted to the problem of defining the variance explained by a set of non-necessarily orthogonal components and selecting one definition.
The properties of the MV formulation of PCA associated to the selected definition are given in Section~\ref{explained variance and block pca}.
Section~\ref{group-sparse block pca formulations} is devoted to the definition, mathematical analysis and algorithmic resolution of the GSMV formulation \eqref{i20}. Finally, numerical results on both synthetic data  and real mixed data are presented in Section~\ref{numerical results}. 

The proposed GSMV algorithm for group sparse PCA and the six explained variance 
definitions of Section~\ref{how to define the explained variance} are implemented in a R package ``sparsePCA'' and are available at \url{https://github.com/chavent/sparsePCA}.

\section{Principal Component Analysis}
\label{pca}

Let $A$ be the data matrix of rank $r$, whose $\ns \times \nvs$ entries  are made of  $\ns$ samples of $\nvs$ centered variables,
 and $\| . \|_{F}$ denote the Frobenius norm on the space of $\ns \times \nvs$ matrices~:
\begin{equation}
\label{105}
  \|A\|_{F}^{2} = \sum_{i=1\dots \ns}\sum_{j=1\dots\nvs} a_{i,j}^{2}
  =\tr(A^{T}A)
  =\sum_{j=1\dots r}\sigma_{j}^{2}\ ,
\end{equation}
where the $\sigma_{j}$'s are the singular values of $A$, defined by its singular value decomposition~:
\begin{equation}
\label{106}\begin{array}{l}
    A=\V\Sigma \W^{T}\quad \mbox{ with } \quad \V^{T}\V=I_{r}\quad,\quad 
    \W^{T}\W=I_{r} \ , \\
 \Sigma=\diag(\sigma_{1}, \dots,\sigma_{r}) = \mbox{$r \times r$ matrix with } \sigma_{1}\geq \sigma_{2}\geq \dots \geq \sigma_{r} >0 \ .
 \end{array}
\end{equation}
The columns $\vv_{1} \dots \vv_{r}$ of $\V$ and  $\ww_{1} \dots \ww_{r}$ of $\W$ are the \emph{left} and \emph{right} singular vectors of $A$.

 Principal Component analysis (PCA) searches for a number $\nl \leq r$ of combinations $z_{j},j=1,\dots \nl$ \emph{(loading vectors)} of the $\nvs$ variables  such that the variables $y_{j}=Az_{j}, j=1 \dots \nl$ \emph{(components)} are uncorrelated and explain an as large as possible fraction of the variance $\|A\|_{F}^{2}$ of the data. The loadings, normalized components and components solution to the PCA  problem are then given by~:
 \begin{equation}
\label{110}
Z^{*} = V_{\nl} \quad , \quad X^{*} = U_{\nl} \quad , \quad Y^{*}= X^{*} \,\diag\{\sigma_{j} ,  j=1 \dots \nl \}
\end{equation}
where the matrices  $Z$ and $Y$ contain the $\nl$ loadings and components, and $U_{\nl}$ and $V_{\nl}$ contain the $\nl$ first left and right singular vectors.
The components $y_{j}^{*}$ are orthogonal, so the part $\varpca$
of the variance of $A$ explained by these $\nl$ components is unambiguously defined by~:
\begin{equation}
\label{115}
  \varpca  = \sum_{j=1 \dots \nl} \|y_{j}^{*}\|^{2} =\|Y^{*}\|_{F}^{2}
  					=\sum_{j=1 \dots \nl} \sigma_{j}^{2}                                  
   					\ \leq \ \sum_{j=1 \dots r} \sigma_{j}^{2}
   					 =  \|A\|_{F}^{2} \ .
\end{equation}

\section{Defining Variance explained by non orthogonal components}
\label{how to define the explained variance}

Definition \eqref{115} for the variance explained by $\nl$ components $Y=AZ$ makes sense as long as the components $y_{j}$ are orthogonal, as it is the case for unconstrained PCA. But sparse PCA algorithms generate usually non orthogonal components, and it is known that the use of (\ref{115}) can lead to overestimate the variance of $Y$, as shown in Section \ref{adjusted optimal and orthogonal variances} below. 
So the problem of \emph{defining} the variance $\var Y$ in that case arises. 

Two definitions have been proposed in the literature. In 2006, \cite{zou2006sparse} introduced the (order dependent) \emph{adjusted variance}, as the sum of the additional variances explained by each new component; in 2008,  \cite{SH2008} introduced an (order independant) {\em total variance}, depending only on the subspace spanned by the components. 
The total variance is bounded by the variance $\|A\|_{F}^{2}$ of $A$, 
\cite{SH2008}[Theorem 1 p.1021], but it is not known wether or not these definitions ensure a diminution of the explained variance with respect to unconstrained PCA, and if they coincide with (\ref{115}) when $Y$ is orthogonal.

So we perform in this section a quite systematic search for possible definitions of the variance $\var Y$, under the constraint that $\var Y$ satisfies a set  of reasonable necessary conditions. This will result in six (including adjusted and total variance) different definitions of $\var Y$ - but we shall end up with only one recommendation.

Let $Y$ be a block of  components associated to a block $Z$ of loadings, with $Y$ and $Z$ linearly independant but possibly non orthogonal~:
\begin{equation}
  Y = AZ  \in \R^{\ns \times \nl}\quad ,\quad Z\in \R^{\nvs \times \nl} 
  \quad , \quad  \rank Y= \rank Z=\nl  \label{500}
\end{equation}
where the number $\nl$ of loadings and components satisfies~:
\begin{equation}
\label{503}
  \nl \leq \rank A \egaldef r \ .
\end{equation}
As it will turn out, the unit norm constraint on the $z_{j}$'s will not always be necessary, so we shall add it only where required.
We want to \emph{define} $\var Y$ in such a way that~:
\begin{itemize}
  \item \textbf{property 1}~: 
  $\var Y$ coincides with $\varpca$ as soon as $Y$ coincides with the PCA components $Y^{*}=AZ^{*}$ recalled in \eqref{110}, that is~:
\begin{equation}
\label{510}
  \var Y^{*}  = \varpca  =  \sum_{j=1 \dots \nl} \sigma_{j}^{2}                                  
   \ \leq \ \sum_{j=1 \dots r} \sigma_{j}^{2}  =  \|A\|_{F}^{2} \quad \text{ for } m=1 \dots r \ .
\end{equation}
  \item \textbf{property 2}~:  for a given number $\nl\leq r$ of unit norm loadings, the variance $\var Y$ is \emph{smaller than the variance obtained by PCA}~: 
\begin{equation}
\label{510a}
\var Y \leq  \varpca = \sigma_{1}^{2} +\dots + \sigma_{\nl}^{2} \ .
\end{equation}
  \item \textbf{property 3}~:  when the components $Y$ happen to be orthogonal, this  variance has to coincide with the usual formula~:
  \begin{equation}
\label{504}
  \var Y = \sum_{j=1\dots \nl} \|y_{j}\|^{2} = \|Y\|_{F}^{2} \ .
\end{equation}
\end{itemize}
We give now two formulas for the computation of $\varpca$ 
which will provide starting points for the definition of $\var Y$ in the case where the components $Y$ are not anymore aligned with the left singular vectors. 

Equation \eqref{115} gives immediately a first formula for $\varpca$~:
\begin{equation}
\label{515a}
\varpca = \sum_{j=1 \dots \nl} \|y_{j}^{*}\|^{2} =\|Y^{*}\|_{F}^{2} \ .
\end{equation}

Then we complement  (\ref{515a}) by an equivalent
 \emph{subspace formulation}   of $\|Y^{*}\|_{F}^{2}$. Let $ \pr_{V_{\nl}}$ denotes the orthogonal projection on the subspace spanned by the $\nl$ first right singular vectors $V_{\nl}=[\ww_{1} \dots \ww_{\nl}]$ of $A$. Then  $\pr_{V_{\nl}} = V_{\nl} V_{\nl}^{T}$, so that $\|A \, \pr_{V_{\nl}}\|_{F}^{2} = \|Y^{*}\|_{F}^{2}$.  This gives a second formula for 
 $\varpca$~:
\begin{equation}
\label{515}
  \varpca =\|A \,\pr_{V_{\nl}}\|_{F}^{2} \ .
\end{equation}
We can now start from either (\ref{515a}) or (\ref{515}) to define the variance  of  the components $Y=AZ$ associated to any block $Z$ of $m\leq r$ linearly independant 
- but not necessarily orthogonal - loading vectors.

 \subsection{Subspace variance}
 \label{section: subspace variance}
We proceed here by analogy with  (\ref{515}), and define, when $Z$ satisfies (\ref{500}), the \emph{subspace variance}  of $Y=AZ$ by~:
\begin{equation}
\label{540}
  \varsubspace Y \egaldef  \|A\pr_{Z}\|_{F}^{2} \ ,
\end{equation}
which shows that  $\varsubspace Y$ coincides with the \emph{total variance} explained by $Y$ introduced by Shen and Huang in \cite[section 2.3 p. 1021]{SH2008}. 

Note that with this definition, $\varsubspace Y$ depends only of the \emph{subspace} spanned by $[z_{1} \dots z_{\nl}]$, so the normalization of loadings $z_{j}$ is not required, as mentioned at the beginning of Section \ref{how to define the explained variance}. Of course, we will still continue to represent loadings by unit norm vectors - but this is here only a convenience.

\begin{Lemma} \emph{(Subspace Variance)}
\label{lem 1}
Let $Z$ satisfy (\ref{500}). Then the \emph{subspace variance} of $Y=AZ$ satisfies~:
\begin{equation}
\label{546}
  \varsubspace \,Y = \tr\big\{Y^{T}Y(Z^{T}Z)^{-1})\big\} \leq \varpca  \ ,
\end{equation}
 and satisfies properties 1 and  2. Moreover~:
\begin{equation}
\label{546-2}
  \varsubspace \,Y = \varpca
   \quad \quad \Leftrightarrow \quad \quad \spann Z = \spann   V_{\nl}   \ .
\end{equation}
When the components  $Y=AZ$ happen to be orthogonal 
, with $\|z_{j}\| =1\, , \, j=1\dots\nl$,
 one has~:
 \begin{equation}
\label{546-1}
  \|Y\|_{F}^{2}  \leq \varsubspace \, Y  \ ,
\end{equation}
and~:
\begin{equation}
\label{546-4}
 \|Y\|_{F}^{2}  = \varsubspace \, Y  \quad \quad \Leftrightarrow \quad \quad
 z_{j} = v_{\ell(j)} \ , \  j=1\dots\nl  \ ,
\end{equation}
where $\ell(j),j=1\dots\nl$ denotes $\nl$  distincts indices among $1\dots r$.
Hence $\varsubspace$ does not satisfy property  3~: 
it will overestimate the variance when the components $Y$ are orthogonal without pointing in the direction of left singular vectors.

\end{Lemma}
The proof is in Section \ref{proof of lem 2} of the Appendix.
Also, when the loadings $Z$ are orthogonal, (\ref{546}) shows that $\varsubspace Y = \|Y\|_{F}^{2}$, but this is again not satisfying as now the components $Y$ are generally not orthogonal.

So we explore in the next section another road in the hope of being able to comply with all properties 1, 2 and 3.

\subsection{Projected  and Normalized variances}
\label{adjusted optimal and orthogonal variances}

We start now from formula (\ref{515a}). A natural generalization
 would be~:
\begin{equation}
\label{520}
  \var Y \egalquest \sum_{j=1 \dots \nl} \|y_{j}\|^{2} =\|Y\|_{F}^{2} = \|AZ\|_{F}^{2}  \ .
\end{equation}
This tentative definition makes sense only if the magnitude of the individual loading vectors if fixed. Hence it has to be used together with the normalization constraint~:
\begin{equation}
\label{522}
 \|z_{j}\|=1 \quad , \quad j = 1 \dots \nl \ .
\end{equation}
In PCA, the loadings $Z$  coincide with right singular vectors
$[ \ww_{1} \dots \ww_{\nl} ]$, and the tentative definition  (\ref{520}) gives~:
\begin{equation}
\label{526}
  \var Y = \|A [\ww_{1} \dots \ww_{\nl}] \|_{F}^{2}   =  \varpca \ ,
\end{equation}
which corresponds to the upper bound required in property 2.
   
In the general case of possibly non orthogonal loadings which satisfy only (\ref{500}) (\ref{522}), property 2 is not ensured anymore with this definition, as many authors have pointed out. For example, consider a matrix $A$ with three singular values $3,2,1$, and chose for $Z$ two linearly independant unit vectors close to the first right singular vector $\ww_{1}$. Then definition (\ref{520}) would give~:
\begin{equation}
\label{528}
  \var Y = \|AZ\|_{F}^{2}= \underbrace{\|Az_{1}\|^{2}}_{\simeq \sigma_{1}^{2}=9} +  \underbrace{\|Az_{2}\|^{2}}_{\simeq \sigma_{1}^{2}=9} \simeq 18 > \underbrace{9+4}_{\sigma_{1}^{2}+\sigma_{2}^{2}}+1=\|A\|_{F}^{2} \ .
\end{equation}
This contradicts both properties 1 and 2, \emph{which
makes (\ref{520}) inadequate as a general definition of variance}.

However, this definition continues to make perfect sense for the variance as long as the components are orthogonal, without pointing necessarily in the direction of left singular vectors~: the components correspond then to a block of independant variables, whose total variance is defined by (\ref{520}). 

Hence a natural way to eliminate the redundancy caused by the orthogonality default of the components $Y$  \emph{and} to satisfy property 3 is to~:
\begin{enumerate}
  \item \textbf{choose a rule to associate to the components $Y$ an \emph{orthonormal basis} $X$ of $\spann\{Y\}$} 
that, loosely speaking, \textbf{ ``points in the direction of the components~$Y$''}~:

\vspace{-3ex}

  \begin{equation}
\label{530}
\begin{array}{c}
     X^{T}X = I_{\nl} \quad, \quad \spann\{X\} = \spann\{Y\} \ ,      \\
     \text{(with the constraint that } x_{j}=y_{j}/\|y_{j}\|  \text{ as soon as }
      \langle y_{j},y_{k}\rangle = 0  \  \forall  j \neq k  \text{ .)}
\end{array}
\end{equation}
We denote by $M$ the matrix of the coordinates of $Y$ in the chosen basis~:\begin{equation}
\label{531}
   M = X^{T}Y \quad \Longleftrightarrow  \quad Y = XM \ .
\end{equation}
Example of such decomposition are the QR and polar decomposition~:
\begin{equation}
\label{533}
  Y = Q\, R \  ,\   Q^{T}Q = I_{\nl} \  ,\ 
  R = \mbox{upper triangular matrix}  \ ,
\end{equation}
\begin{equation}
\label{535}
   Y = U\, P \  ,\   U^{T}U = I_{\nl} \   ,\  
   P^{T} \!= P \in \R^{\nl \times \nl} \  , \  P \geq 0  \ .
\end{equation}
\item \textbf{associate to $Y$ \emph{orthogonal modified components} $Y\p$ along the $X$ axes, and \emph{define} the variance $\var Y$ explained by the components $Y$ by~:}
\begin{equation}
\label{533-1}
  \var Y \egaldef \|Y\p\|_{F}^{2} \ .
\end{equation}  
\end{enumerate}

\subsubsection{Projected Variance}
\label{section: projected explained variance}
We define here the modified components $y\p_{j}$ as the projection of $y_{j}$ on the  $j$-th axis of the basis 
  $X$~:
\begin{equation}
\label{532}
  {y\p}_{j} =  \langle y_{j}\,,x_{j}\rangle x_{j} = m_{j,j}\,x_{j} \quad , \quad j=1\dots\nl \ .
\end{equation}  

\vspace{-1.5ex}

\begin{Lemma} \emph{(Projected variance)}
\label{lem 3}
For any choice of \emph{orthogonal basis} $X$ satisfying (\ref{530}), the
 variance $\varproj Y$ defined by \eqref{533-1} and (\ref{532})  satisfies~:
\begin{equation}
\label{534-2}
     \varproj Y = \sum_{j=1\dots\nl} \langle y_{j}\,,x_{j}\rangle^{2} =  \tr\{\diag^{2}M\} 
  \leq \varsubspace \, Y
  \leq \varpca  \ .
\end{equation}
and  Properties 1, 2 and 3 are satisfied. Moreover,
\begin{equation}
\label{534-6}
\exists X \ s.t. \  \varproj Y =   \varpca \quad\quad \Leftrightarrow 
 \quad \quad 
  \left\{\begin{array}{lcl}
    \spann Z &=& \spann   V_{\nl}  \ , \\
z_{j}^{T}V_{\nl}\Sigma_{\nl}^{-2}V_{\nl}^{T}z_{k} & = & 0   \text{ for } j \neq k
\end{array} \right.
\end{equation}
 Of course, $z_{j}=v_{j},j=1\dots\nl$ satisfies the right part of \eqref{534-6}~!
\end{Lemma}
The proof is given in Section \ref{proof of lem 3} of the Appendix. 
Depending on the rule chosen  at step 1  for the selection of $X$, formula (\ref{534-2}) gives different possible definitions for the variance, which 
all satisfy  Lemmas~\ref{lem 3}~:

\vspace{1ex}

\noindent\textbf{Adjusted variance.} Let $X$ and $M$ be given by the  \emph{QR-decomposition} (\ref{533}) of $Y$ ($X=Q,M=R$).  Then (\ref{534-2}) gives~:
\begin{equation}
\label{534}
 \varprojqr Y =   \tr \{ \diag^{2} R \}
                 = \tr\{ R^{2} \} = \langle R^{T},R \rangle_{F} \ ,
\end{equation}
which is the \emph{adjusted variance} introduced by Zou et al. in \cite{zou2006sparse}.
Because the QR orthogonalization procedure is started with the components of largest norm, the  basis $X=Q$ will point in the direction of $Y$ at least for the components of larger norm.

\vspace{1ex}

\noindent\textbf{Polar variance.} Define  $X$ and $M$ by the \emph{polar decomposition} $UP$ of $Y$ (\ref{535}). The choice $X=U$
does its best to point in the same direction as the components $Y$, in that it maximizes
the scalar product $\langle Y,X \rangle$ over $\stns$.
Formula  (\ref{534-2})   gives now~:
\begin{equation}
\label{536}
    \varprojup Y      =   \tr\{ \diag^{2} P\} = \tr\{ (\diag^{2} (Y^{T}Y)^{1/2} \} \ .
\end{equation}
This  \emph{polar variance} is order independant.

\vspace{1ex}

\noindent\textbf{Optimal variance.} Lemmas \ref{lem 3} shows that the variance $\varproj Y$ defined by    (\ref{534-2}) satisfies the desired properties 1,2 and 3 for any $X$ such that 
$X^{T}X = I_{\nl}$. It is hence natural to associate to $Y$ the  basis $X^{*}$ which gives the {largest variance}~:
\begin{equation}
\label{534-10}
 \varprojopt Y =  \sum_{j=1\dots\nl} \langle y_{j}\,,x_{j}^{*}\rangle^{2}
 = \max_{X\in\stns} \sum_{j=1\dots\nl} \langle y_{j}\,,x_{j}\rangle^{2} \ ,
\end{equation}
where $\stns$ is the \emph{Stiefel variety} of the component space~:
\begin{equation}
\label{135}
  \stns=\{X \in \R^{\ns \times \nl} \mbox{ such that } 
  X^{T}X = I_{\nl}   \}   \ .
\end{equation}
This optimal variance can be computed by maximization
of the convex function $\varproj Y$ over the Stiefel manifold $\stns$ by mean of \cite[Algorithm 1 page 526]{JNRS2010} recalled in  \eqref{414-1}, which gives~:
\begin{equation}
\label{534-11}
  X^{*}= \lim_{k\rightarrow \infty} X_{k} \quad \text{ where: }\quad
  X_{k+1} = \polar \big( Y \diag(X_{k}^{T}Y)    \big) \quad , \quad
  X_{0}=U=\polar(Y) \ .
\end{equation}
This shows that $X^{*}$ is a solution of the equation~:
\begin{equation}
\label{534-14}
X = \polar \big( Y \diag(X^{T}Y)) \ .
\end{equation}

\vspace{-1,5ex}

\subsubsection{Normalized Variance}
We choose now ${y\p}_{j}$ in the direction of $x_{j}$ such that~:
 \begin{equation}
\label{532-1}
   {y\p}_{j} = A {z\p}_{j} \quad , \quad \mbox{with}  \quad \|z\p_{j}\| = 1 \ , \  j=1\dots\nl \ ,
   \quad \text{where~:}
\end{equation}
\begin{equation}
\label{533-2}
 z\p_{j} = \frac{t_{j}}{\|t_{j}\|} \ , \text{ with $T$ is given by~: }
  Z = T M  \quad  \text{(compare to (\ref{531})~: $Y=XM$)} \ .
\end{equation}
 One sees that $T$ is obtained by performing on the loadings $Z$ the same linear combinations  which transformed $Y$ into $X$, so that $AT=X$.
\begin{Lemma} \emph{(Normalized variance)}
\label{lem 4}
For any choice of \emph{orthogonal basis} $X$ satisfying (\ref{530}), the 
variance $\varnorm Y$  defined by  \eqref{533-1} and  \eqref{532-1} \eqref{533-2} satisfies~:
\begin{equation}
\label{534-3}
     \varnorm Y = \sum_{j=1\dots\nl} 1/\|{t}_{j}\|^{2}
     \leq \varsubspace \, Y  \leq  \varpca  \ ,
\end{equation}
and satisfies Properties 1,  2 and 3. Moreover~:
\begin{equation}
\label{534-8}
 \exists X \  s.t. \ \varnorm Y =   \varpca 
 \quad \quad \Leftrightarrow \quad \quad \spann Z = \spann V_{\nl}
\end{equation}
\end{Lemma}
The proof of \eqref{534-3} follows immediately from \eqref{546-1} in Lemma \ref{lem 1} applied to the orthogonal components $Y\p=AZ\p$. In order to prove 
\eqref{534-8}, one remarks first that $\var_{norm}Y = \varpca$ implies by \eqref{534-3}
that $\varsubspace Y=\varpca$, which in turn implies by Lemma \ref{lem 1} that $\spann Z = \spann V_{\nl}$. Conversely, if  $\spann Z = \spann V_{\nl}$, one can choose for $X$ the $\nl$ first left singular vectors $U_{\nl}$, which gives 
$\varnorm Y= \|Y \p \|^{2} = \varpca$, which ends the proof of the lemma.

\vspace{1,5ex}

\noindent\textbf{QR normalized variances.} Let  $X$ and $M$ be defined by the
 \emph{QR decomposition} (\ref{533}) of $Y$. Then (\ref{534-3}) leads to another definition of variance~:
\begin{equation}
\label{539-0}
  \varnormqr Y= \sum_{j=1 \dots \nl} 1/\|t_{j}\|^{2} 
  = \tr\{ \diag^{-1}(T^{T}T)  \}  \quad \mbox{ where } \quad T=ZR^{-1}  \ .
  \end{equation}
\noindent\textbf{UP normalized variances.} Let $X$ and $M$ by defined by the  $UP$ \emph{polar decomposition}  (\ref{535}) of $Y$. Then 
 (\ref{534-3}) defines a new variance~:
\begin{equation}
\label{539-1}
  \varnormup Y\!=\!\!  \sum_{j=1 \dots \nl} \!\!   1/\|t_{j}\|^{2} 
  = \tr\{ \diag^{-1}(T^{T}T)  \}   \ \mbox{ where } \ T=Z(Y^{T}Y)^{-1/2} \ . 
  \hspace{-2em}
  \end{equation}

 We have not considered defining an \emph{optimal normalized variance} by associating to $Y$ the basis  $X$ which maximizes  $\var Y$ defined by \eqref{534-3}
 \eqref{533-2}, for the reason that, because of  \eqref{534-8}, such an optimal normalized variance would achieve its maximum $\varpca$ as soon as $\spann Z = \spann V_{\nl}$,  no orthogonality condition being required on $Z$ as it was the case for projected variance. 
 Another reason for this choice is that $\var Y$ is not anymore a convex function of $X$, so its computation would be more delicate.

\subsection{Size comparison}
\label{comparison of explained variances}

The variance $\varsubspace Y$ is larger than any of the five other variances, as shown by Lemmas  \ref{lem 3} and \ref{lem 4}. 

\begin{figure}[h]
\begin{center}
\centerline{\resizebox{28em}{!}{\includegraphics{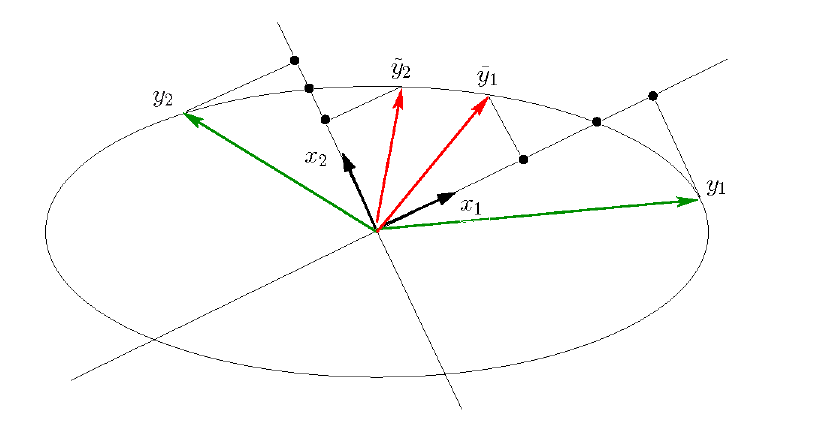}}}
\caption{The two sets of components $Y=[y_{1}\,y_{2}]$ and $\tilde{Y}=[\tilde{y}_{1}\,\tilde{y}_{2}]$ have been chosen such that their polar decomposition produces the same basis $X=[x_{1}\,x_{2}]$, and one sees that~:
 $ \varprojup \, \tilde{Y} \leq \varnormup \tilde{Y} = \varnormup Y \leq  \varprojup Y $.}
\label{fig 1}
\end{center}
\end{figure}

But there is no natural ordering between the variances defined by projection and normalization, as illustrated in Figure  \ref{fig 1} for the case of polar decomposition.

Then among projected variances, 
$\varprojopt Y$ is greater by definition than  any other projected variance, in particular greater than $\varprojqr Y$ and $\varprojup Y$. 
But there is no natural order between these latter~: when the components $Y$ are of equal norm, one checks easily 
 that the basis $X$ which maximizes the variance is $\polar Y$, which implies in particular that~:
 \begin{equation}
\label{539-27}
\varprojopt Y =\varprojup Y \geq \varprojqr Y \ .
\end{equation}
But the converse of the last inequality can hold when
 the norms of the components are very different~: for $\nl = 2$, one checks that 
$\|y_{2}\|/\|y_{1}\|$ small enough implies that $\varprojup Y \leq \varprojqr Y$.

 There is also no natural order between the normalized variances $\varnormqr Y$ and $\varnormup Y$~: for components $Y$ such that the basis $X$ associated by QR-decomposition coincides with the $\nl$-first left singular vectors $U_{\nl}$ of $A$, one has, according to Lemma \ref{lem 4}~:
 \begin{equation}
\label{539-3}
  \varnormqr \,Y = \varpca \geq \varnormup \, Y \ ,
\end{equation}
  with a strict inequality as soon as $Y$ and $U_{\nl}\diag\{ \sigma_{j}\}$ don't coincide.
 The same reasoning with the polar decomposition in place of the QR decomposition shows that the converse inequality can happen.
 
We   complete now these theoretical results by comparing numerically  the six definitions of $\var Y$ proposed Section \ref{how to define the explained variance} and summarized in Table \ref{res_def_var}.

\begin{table}[ht]
\centering
\begin{tabular}{lll}
  \hline
 Name & Notation & Short name \\ 
  \hline
subspace variance & $ \varsubspace$ & subspVar \\ 
 optimal variance &  $\varprojopt $ & optVar \\ 
  polar variance & $\varprojup$  & polVar \\ 
  adjusted variance &  $\varprojqr$  & adjVar  \\ 
 QR normalized variance & $\varnormqr$ & QRnormVar  \\ 
  UP normalized variance & $\varnormup$ & UPnormVar \\ 
   \hline
\end{tabular}
\captionof{table}{Names and notation for the 6 variance definitions.}
\label{res_def_var}

\end{table}

The numerical comparison of the six definitions  of variance is made on
sets of $30 000$ non orthogonal components $Y$ computed as follows:
\begin{itemize}
\item [-]Two sets of $100$ matrices $A$ each are simulated using the simulation scheme of Section \ref{num res simulated data}. The matrices of the first set  (referred to as ``close eigenvalues'') are drawn randomly using a covariance matrix whose first eigenvalues  are  200, 180, 150 and 130. The matrices of the second set (referred to as ``different eigenvalues'')  are   drawn randomly using a covariance matrix whose first eigenvalues  are 200, 100, 50 and 20.
\item[-] Three group-sparse PCA algorithms (one deflation and two block algorithms described in Section \ref{block vs deflation} ) are applied to these simulated matrices with a grid of 50 reduced sparsity parameter $\lambda$ (see \eqref{nr5} in Section \ref{choice reg par}) between 0 and 0.5. For each sparsity parameter $\lambda$, a set of 300 matrices  $Z$  (of $m=4$ loading vectors) are then obtained with the  ``close eigenvalues''  (resp. ``different eigenvalues'' ) set of matrices $A$. 
\item[-] For each loading matrix $Z$,  the six variance definitions of $\var Y=\var(AZ)$ are computed
\end{itemize}

We display here for each variance definition the  {dimensionless} {\em proportion of explained variance} (pev) defined by~:
\begin{equation}
\label{nr25}
  0 \leq  \mathrm{pev} = \var Y\big  / \|A\|_{F}^{2} \leq
   \varpca \big / \|A\|_{F}^{2}  \ ,
\end{equation}
where the right inequality follows from \eqref{510a} in Property 2, which is satisfied by all definitions of variance. Figure \ref{fig 8} gives thus for each definition of $\var Y$ the  {\em mean values} of the pev  over the 300 non orthogonal components $Y=AZ$ (for the ``close eigenvalues'' and `the `different eigenvalues'' case) as a function of the reduced sparsity parameter $\lambda$.

For the ``close eigenvalues '' case (top), where the variances of the principal components are in a ratio less than 1:2, the mean pev's produced by optVar and
 polVar are so close that they cannot be distinguished on the figure (remember that they would coincide were the norms equal, see \eqref{539-27}). 
One sees also that  the {\em mean pev's} seem to be in an apparent 
order for all~$\lambda$~:
\begin{equation}
\label{nr 30}
\text{subspVar  $\geq$ optVar $\geq$ polVar $\geq$ 
adjVar $\geq$ QRnormVar $\geq$ UPnormVar}
\end{equation}
(the two first inequalities are not a surprise, as they hold already for any realization 
$Y$ of the components according to Lemma \ref{lem 3} and the definition of 
$\varprojopt$).

But in the ``different eigenvalues'' case (bottom), where the ratio of the principal components variances is 1:10, a zoom on the curves shows that 
optVar is clearly larger than polVar
 and adjVar (which is coherent with the theoretical results), but that the apparent order \eqref{nr 30} is not satisfied  by the other mean pev's anymore.
  
\noindent
\begin{minipage}[c]{1\textwidth}
\centering
\resizebox{24em}{!}{\includegraphics{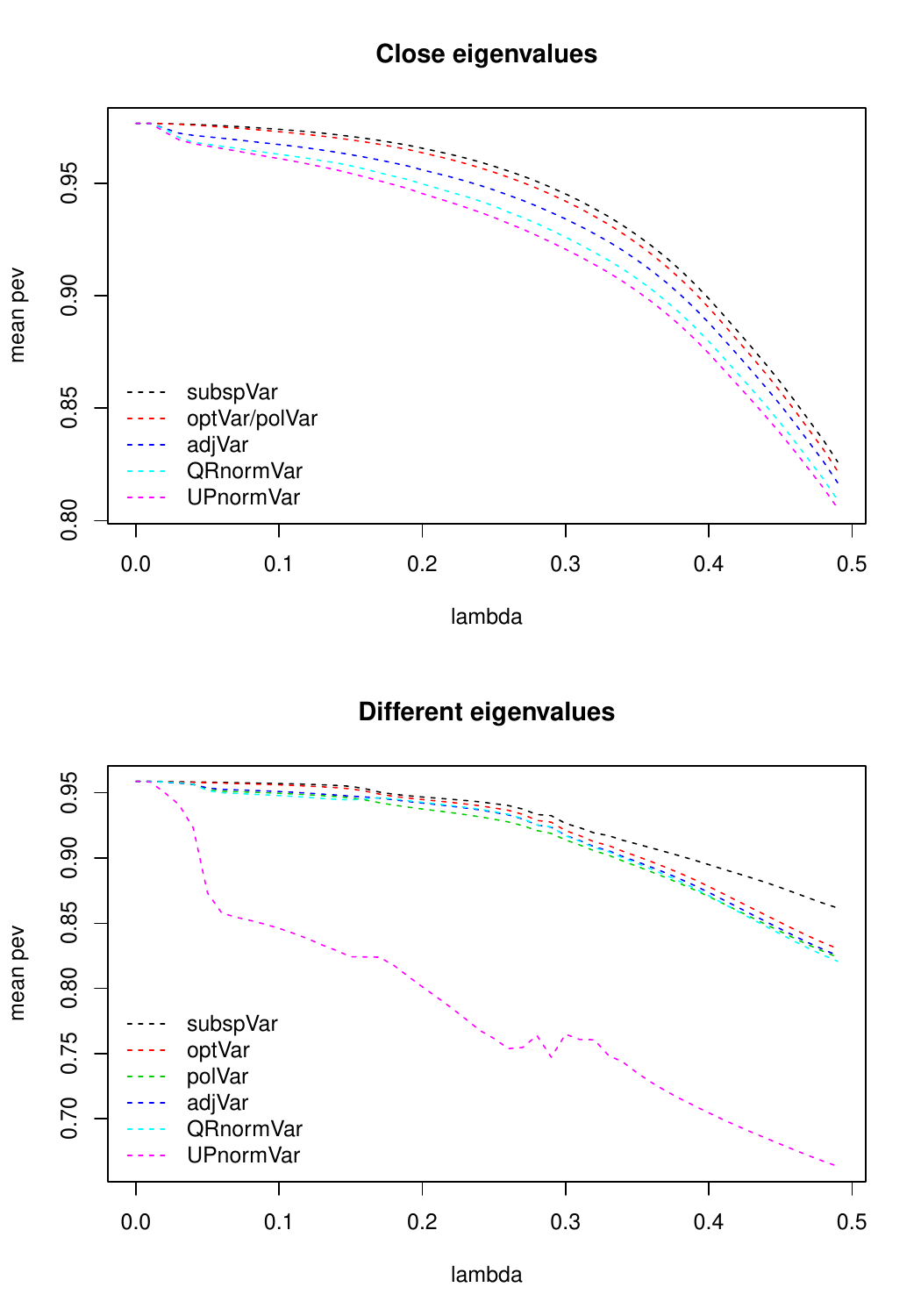}}
\captionof{figure}{Comparison of the  mean pev's (proportion of explained variance) over two sets of components (top and bottom) as function of reduced sparsity parameter $\lambda$ 
(see \eqref{nr5} in Section \ref{resolution of the group-sparse component/loading block formulation})
for the six variance definitions.}
\label{fig 8}
\end{minipage}

\vspace{2ex}

When it comes to real data, the variability of  variance is an important feature, as only one realization is available.  
Table \ref{tab_sd_var} shows that all definitions but UPnormVar (in the "Different eigenvalues " case) exhibit quite similar dispersions.

\begin{table}[ht]
\centering
\begin{tabular}{rrr}
  \hline
 & Close eigenvalues & Different eigenvalues \\ 
  \hline
subspVar & 0.63 & 1.63 \\ 
  optVar & 0.66 & 1.38 \\ 
  polVar & 0.66 & 1.06 \\ 
  adjVar & 0.72 & 1.25 \\ 
  QRnormVar & 1.06 & 1.33 \\ 
  UPnormVar & 1.21 & 7.74 \\ 
   \hline
\end{tabular}
\captionof{table}{Standard deviations$\times 100$  of the six pev (proportion of explained variance) obtained for $\lambda=0.3$ with the three algorithms over the two sets of components (close eigenvalues and different eigenvalues).}
\label{tab_sd_var}

\end{table}

\subsection{Ranking properties of variances}
\label{ranking properties of explained variances}

The proportions of explained variance $\pev_{i},i = 1 \dots 6$ defined by (\ref{nr25}) are meant to be used for the ranking of algorithms, so it is important to figure out wether or not definitions $i$ and $j$  will rank in the same order the components $Y$ and $Y^{\prime}$ obtained from possibly different algorithms and/or sparsity parameter $\lambda$ and/or realization of the data matrix $A$. There are 3 algorithms, 50 values of $\lambda$ and 100 realizations of $A$ (`close eigenvalues '' case), and hence $15000\times14999/2$ couples $(Y,Y^{\prime})$ to be tested. Among these couples, we may consider as  
$\epsilon$-distinguishable from the point of view of our explained variances those for which
\begin{equation}
\label{nr30}
 |\pev_{i}(Y)-\pev_{i}(Y^{\prime})| \geq \epsilon  \quad  \mbox{ for all } \quad i=1 \dots 6 
\end{equation}
for some $\epsilon\geq 0$. Table \ref{tab 2} shows the percentage of cases where $\pev_{i}$ and $\pev_{j}$ rank identically components $Y$ and $Y^{\prime}$ among all $\epsilon$-distinguishable couples. The good news is that all three projected variances optVar, polVar and adjVar, as well as the normalized variance QRnormVar,  produce the same ranking as soon as one considers that differences in proportion of explained variance  under $10^{-2}$ are not significative. For the same $\epsilon$, the two other definitions subspVar and UPnormVar still produce quite different rankings. Of course, this is only an experimental result based on our simulated data sets.

\begin{table}[ht]
\centering
\begin{tabular}{rrrrrrr}
  \hline
 & subspVar & optVar & polVar & adjVar & QRnormVar & UPnormVar \\ 
  \hline
subspVar &  & 79.98 & 71.05 & 70.71 & 69.75 & 56.19 \\ 
  optVar &  &  & 88.93 & 89.87 & 89.19 & 73.71 \\ 
  polVar &  &  &  & 96.22 & 95.27 & 84.62 \\ 
  adjVar &  &  &  &  & 98.34 & 83.15 \\ 
  QRnormVar &  &  &  &  &  & 82.75 \\ 
  UPnormVar &  &  &  &  &  &  \\ 
   \hline
\end{tabular}

\vspace{2ex}

\centering
\begin{tabular}{rrrrrrr}
  \hline
 & subspVar & optVar & polVar & adjVar & QRnormVar & UPnormVar \\ 
  \hline
subspVar &  & 86.13 & 83.78 & 84.70 & 84.57 & 68.68 \\ 
  optVar &  &  & 96.40 & 98.57 & 98.32 & 81.14 \\ 
  polVar &  &  &  & 97.81 & 97.72 & 84.74 \\ 
  adjVar &  &  &  &  & 99.66 & 82.55 \\ 
  QRnormVar &  &  &  &  &  & 82.66 \\ 
  UPnormVar &  &  &  &  &  &  \\ 
   \hline
\end{tabular}

\vspace{2ex}

\centering
\begin{tabular}{rrrrrrr}
  \hline
 & subspVar & optVar & polVar & adjVar & QRnormVar & UPnormVar \\ 
  \hline
subspVar &  & 89.57 & 89.57 & 89.57 & 89.57 & 68.80 \\ 
  optVar &  &  & 100.00 & 100.00 & 100.00 & 79.23 \\ 
  polVar &  &  &  & 100.00 & 100.00 & 79.23 \\ 
  adjVar &  &  &  &  & 100.00 & 79.23 \\ 
  QRnormVar &  &  &  &  &  & 79.23 \\ 
  UPnormVar &  &  &  &  &  &  \\ 
   \hline
\end{tabular}

\captionof{table}{The  entry of each table on line $i$ and  column $j$ gives the percentage of $\epsilon$-distinguishable couples $Y,Y^{\prime}$ which are ranked identically by $\pev_{i}$ and $\pev_{j}$. Top~: $\epsilon=0$, middle~: $\epsilon=10^{-3}$, bottom~: $\epsilon=10^{-2}$.}
\label{tab 2}
\end{table}

\subsection{Recommendation for the choice of a definition of variance}
\label{subsec: which explained variance do we recommend}

Among the definitions of variance we have investigated,   the subspace variance $\varsubspace Y$ (Lemma \ref{lem 1}) fails to satisfy property 3~: when the components $Y$ happen to be orthogonal without pointing in the directions of left singular vectors, $\varsubspace Y > \|Y\|_{F}^{2}$, where one would like equality. So $\varsubspace Y$ does not seem to be the best choice.

Then Lemma \ref{lem 3} and  \ref{lem 4} show that all projected and normalized  variances satisfy properties 1, 2 and 3. We rank them now according to their ability to achieve their maximum value $\varpca$~:  the more restrictive the conditions on the loadings $Z$, the lesser the chance that 
$\var Y = \varpca$~!
This is a desirable property, as it will allow to quantify the drop in explained variance with respect to PCA induced by using sparse loading, and will help to
make a decision in the trade-off ``explained variance versus sparsity''.

For normalized variances, the possibility that $\varnorm Y$ achieves the PCA value $\varpca$ will arise as soon as $\spann Z = \spann   V_{\nl}$ (Lemma  \ref{lem 4}), if the basis $X$ associated to $Y$ happens to be the basis $U_{\nl}$ of left singular vectors. Hence equality can hold for a large family of components, so $\varnorm Y$ does not seem either to be the best choice.

For the projected variances $\varproj Y$,  achieving the PCA maximum $\varpca$ requires not only that $\spann Z = \spann   V_{\nl}$,  
\emph{but also that}  the loadings $Z$
satisfy a weighted orthogonality condition (Lemma \ref{lem 3}). 
They seem to be the best suited to arbitrate the trade-off between explaining the variance and achieving sparsity of the loadings, as $\varproj Y<\varpca$ as soon as the loadings fail to span the $\nl$ first right singular vector \emph{or} to satisfy the weighted orthogonality condition. So projected explained variances will be our preferred definitions.

In short, for the projected variances,  $\varprojqr$ is easy to compute but  order dependent, $\varprojup$ is order independant, and $\varprojopt$ is larger than any projected variance (including $\varprojqr$ and $\varprojup$), but its computation requires an iterated polar decomposition. They are all good choices, our best choice for variance being $\varprojopt$, which we shall use in the sequel as \emph{the} definition of variance, and hence drop the indices ``proj'' and ``opt''~:

\begin{Definition}
\label{def 1}
The \emph{optimal projected variance} of $\nl$ not necessarily orthogonal components $Y=[y_{1} \dots y_{\nl}]$ is~:
\begin{equation}
\label{370}
\var Y =  \max_{X\in\stns} \sum_{j=1\dots\nl} \langle y_{j}\,,x_{j}\rangle^{2} \ .
\end{equation}
This definition satisfies properties 1, 2 and 3, in particular  it reduces to the usual formula $\var Y = \|Y\|_{F}^{2}$ when  the components $Y$ are orthogonal, and is larger than all projected variance definitions.
\end{Definition}

\section{Maximum Variance (MV) Block PCA formulations}
\label{explained variance and block pca}

Because of properties \eqref{534-2} \eqref{534-6} of Lemma \ref{lem 3}, the optimal projected variance $\var Y$ of Definition \ref{def 1} achieves its maximum $\varpca=\sigma_{1}^{2}+
\dots+\sigma_{\nl}^{2}$ for any loading $Z^{*}$ which satisfies  \eqref{534-6}, in other terms~:
\begin{equation}
\label{132}
Z^{*} = \arg \max_{Z \in \bpnvs} \var(AZ) \quad
\Leftrightarrow  \quad
 \left\{\begin{array}{lcl}
    \spann Z^{*} &=& \spann   V_{\nl}  \ , \\
     z_{j}^{*T}V_{\nl}\Sigma_{\nl}^{-2}V_{\nl}^{T}z_{k}^{*} & = & 0   \text{ for } j \neq k
\end{array} \right.
  \ ,
\end{equation}
where~:
\begin{equation}
\label{134}
  \bpnvs=\{Z \in \R^{\nvs \times \nl} \mbox{ such that } 
  \|z_{j}\|\leq1\ , \ j=1 \dots \nl \}   \ .
\end{equation}
So the variance $\var Y$  can be used as an objective function whose maximization with respect to $\nl$ (non-necessarily orthogonal) unit norm loadings produces 
$\nl$ (non-necessarily orthogonal) components $Y^{*}=AZ^{*}$ with variance 
$\varpca=\sigma_{1}^{2}+\dots+\sigma_{\nl}^{2}$.
Note that the blocks $Z^{*}=V_{\nl}$ made of the $\nl$ first right singular vector, and all those obtained by column index permutation, satisfy trivially the property in the right part of \eqref{132} and hence are all maximizers of $\var(AZ)$

In order to select the maximizer $Z^{*}=V_{\nl}$ of interest for PCA,
one introduces weights $\mu_{j}$ such that~:
\begin{equation}
\label{136}
  \mu_{1} \geq  \mu_{2} \geq \dots  \mu_{\nl} >0 \ ,
 \end{equation}
and defines a  ``weighted variance'' $\varmu$ by~:
 \begin{equation}
\label{136-4}
  \varmu (AZ)=   \max_{X\in\stns} \sum_{j=1\dots\nl} \mu_{j}^{2} \langle Az_{j}\,,x_{j}\rangle^{2} \ ,
\end{equation}
which coincides with $\var(AZ)$ when $\mu_{j}=1 \text{ for all } j$.
The \emph{Maximum Variance (MV)} Block PCA formulation is then defined as~:
\begin{equation} 
\label{132-4}
\max_{Z \in \bpnvs} \varmu(AZ)  =   
 \max_{Z \in \bpnvs}   \max_{X\in\stns} \sum_{j=1\dots\nl} \mu_{j}^{2} \langle Az_{j}\,,x_{j}\rangle^{2}  =
  \max_{X\in\stns}  \sum_{j=1\dots\nl} \mu_{j}^{2} \|A^{T}x_{j}\|^{2} \ .
\end{equation}
The nice properties  of this formulation are recalled in the next proposition~:
\begin{Proposition}
\label{pro 1-Z}
Let the singular values of $A$ satisfy~:
\begin{equation}
\label{137}
  \sigma_{1}>\sigma_{2}> \dots >\sigma_{\nl}>0 \ ,
\end{equation}
and the weights $\mu_{j}$ satisfy (\ref{136}).  Then the PCA loadings $Z^{*}=V_{\nl}$ and normalized components $X^{*}=U_{\nl}$ defined in  \eqref{110} are \emph{one} solution of the MV problem  \eqref{132-4} when the weights $\mu_{j}$ are constant, and \emph{the unique} solution (up to a multiplication by $\pm1$ of each column of course) of the MV problem \eqref{132-4} when the weights $\mu_{j}$ are strictly decreasing.
The maximizers $Z^{*}$ and $X^{*}$ are hence independant of the weights $\mu_{j}$, and the variance explained by $Y^{*}=AZ^{*} $ is~: 
\begin{equation}
\label{147a}   
   \var \, Y^{*}   =  \varpca =  \sum_{j=1\dots\nl} \sigma_{j}^{2} \leq 
   \|A\|_{F}^{2} \ ,
\end{equation}
with $X^{*}$ and $Z^{*}$ related by~:
 \begin{equation}
\label{162}
  x_{j}^{*}=(Az_{j}^{*}) / \|Az_{j}^{*}\| \quad , \quad  z_{j}^{*}=(A^{T}x_{j}^{*}) / \|A^{T}x_{j}^{*}\| 
  \quad \mbox{for} \quad j=1 \dots \nl \ .
\end{equation}
\end{Proposition}
\noindent\textbf{Proof:}
The last formulation in resp. \eqref{132-4}
is the maximization of a weighted Rayleigh quotient
for $A^{T}$, which is known to be equivalent to a PCA problem for $A^{T}$, and hence for $A$ (see for example \cite{AMS2008}, recalled as Theorem~\ref{thm GRQ} in the Appendix  for the case of constant weights, and 
\cite{Brockett1991} for  the case of decreasing weights).
\cqfd

\section{Group-Sparse Maximum Variance (GSMV) Block PCA formulation}
\label{group-sparse block pca formulations}

 The aim of group sparse PCA is to build group sparse loading vectors $z_{j}$ in order to select relevant groups of variables to build the component $y_{j}$. This is helpful in particular for the treatment of mixed data, where a group of scalar variables is used to describe one categorical variable.

We denote by $\nvg$  the number of groups of variables (also called ``group variables'' below) and by
$\dv_{1}, \dots ,\dv_{\nvg}$ the number of variables in each group.
The data matrix $A$ and the loading vectors $z_{j}\in \R^{\nvs}, j=1\dots \nl$ are then split accordingly~:
\begin{equation}
\label{301}
  A= \big[ A_{1} \dots A_{\nvg} \big] \quad , \quad 
   z_{j}^{T}= \big[ z_{1,j}^{T} \dots z_{\nvg,j}^{T} \big] \ ,
\end{equation}
where the $A_{i}$'s are $\ns\times\dv_{i}$ matrices, and
 the $z_{i,j}$'s are column vectors of dimension $\dv_{i}$. 
 We denote also by 
$\| . \|_{2}$ the norm on $\ns\times\dv_{i}$ matrices induced by the Euclidian norms $\| . \|$ on $\R^{\ns}$ and $\R^{\dv_{i}}$ (largest singular value)~:
\begin{equation}
\label{303}
 \|A_{i} z_{i,j}\| \leq \|A_{i}\|_{2} \|z_{i,j}\| \quad \forall  z_{i,j} \in \R^{\dv_{i}} \ .
\end{equation}

\subsection{Choice of the formulation}
\label{choice of the formulation}

 We start from the MV Block PCA formulation  \eqref{132-4} of Section  section~\ref{explained variance and block pca}, whose unique solution, for strictly decreasing $\mu_{j}$, happens to maximize also $\var(AZ)$ (Proposition \ref{pro 1-Z}).
In order to promote the apparition of zeroes in the loading vectors for some group variables, we use the {\em group $\ell^{1}$-norm} of the  loadings $z_{j}$~:
\begin{equation}
\label{304}
\|z_{j}\|_{1}=\sum_{i=1}^{\nvg} \|z_{i,j}\| \quad , \quad j=1\dots\nl  \ ,
\end{equation}
where $\|z_{i,j}\|$ is the Euclidean norm on $\R^{\dv_{i}}$, and choose regularization parameters~:
\begin{equation}
\label{305}
  \gamma_{j}>0 \ ,\ j=1\dots\nl \ .
\end{equation}
We use this group ${\ell^{1}}$-norm to define a  group-sparse  weighted variance $\varmugamma$ associated to the (non-necessarily orthogonal) loadings $Z \in \bpnvs$, which penalizes loadings  $Z$ whose columns $z_{j}$ have large group $\ell^{1}$-norm (compare with \eqref{136-4})~:
\begin{equation}
\label{305-3}
  \varmugamma (AZ) = \max_{X \in \stns} \sum_{j=1\dots\nl}\mu_{j}^{2}
   \big[ x_{j}^{T}Az_{j} -\gamma_{j}\|z_{j}\|_{1} \big]_{+}^{2} \ ,
\end{equation}
where $[t]_{+}=t$ if $t\ge0$ and $[t]_{+}=0$ if $t<0$. 
Maximization of this variance  leads to the  \emph{Group-Sparse Maximum Variance} (GSMV) Block PCA formulation~: 
\begin{equation}
\label{350}
     \max_{Z \in \bpnvs}  \varmugamma(AZ) =  \max_{Z \in \bpnvs}  \max_{X \in \stns}  \sum_{j=1\dots\nl}\mu_{j}^{2}
   \big[ x_{j}^{T}Az_{j} -\gamma_{j}\|z_{j}\|_{1} \big]_{+}^{2} \ .
\end{equation}
It  produces \emph{non-necessarily orthonormal} sparse loading vectors $z_{j}^{*}$, and orthonormal vectors $x_{j}^{*}$ - but these latter do not coincide anymore with the normalized component~:
\begin{equation}
\label{360}
  x_{j}^{*} \neq (Az_{j}^{*})/\|Az_{j}^{*}\| \ , \  j=1 \dots \nl \ ,
\end{equation}
in opposition to the case where no sparsity is required, where equality holds (see  (\ref{162})).
Hence neither the sparse loading vectors nor the principal components produced by formulation \eqref{350} are orthogonal.

But the good side of this formulation is that the numerical difficulties are split between $X$ and $Z$~: the orthonormality constraint is for $X$, the non-differentiable group $\ell^{1}$-norm is for $Z$.
Moreover, as it was proved, for scalar variables, by \cite{d2008optimal} in the case of cardinality regularization, and indicated  by  \cite{JNRS2010} in the case of $\ell^{1}$ regularization,  the maximization loop on $Z$ in (\ref{350}) can be solved analytically, despite the non-differentiable terms, for any given $X \in \stns$, as it was trivially the case in \eqref{132-4} for PCA,  thus leading to the \emph{maximization of the differentiable convex} function $\Fmugamma$ of $X$ given by \eqref{404} below (to be compared to the last maximization problem  in  \eqref{132-4}),
which  coincides, for scalar variables,   with the function 
$\Phi_{\ell_{1},m}^{2}$ indicated in  \cite[formula (16) page 524]{JNRS2010}, which is the starting point of its $\ell_{1}-$penalty method.

Hence formulation  \eqref{350} provides a natural generalization of this algorithm to the case of group variables, together with
a new interpretation of the algorithm as a sparse version of the maximization of the explained variance $\var(AZ)$ over all unit norm loadings $Z$~: it realizes a compromise between maximizing the variance $\var Y$ explained by $Y$ and sparsifying the loadings $Z$.

We detail now the resolution of the GSMV formulation \eqref{350}.
 For any $j=1 \dots \nl$ we denote by $S_{j}$ the \emph{soft group thresholding operator} 
 which acts on any set of group variables $\zeta=(\zeta_{1}\dots\zeta_{\nvg})\in \R^{\nvs}$ by setting, for each $i=1\dots\nvg$,
 the $i$-th group of variables $\|\zeta_{i}\|$ to zero if its norm is smaller than 
 $\gamma_{j}$, and by reducing its length by $\gamma_{j}$ otherwise~:
 \begin{equation}
\label{402}
\big( S_{j}(\zeta) \big)_{i} = 
\left\{ \begin{array}{lll}
   \zeta_{i} (1-\eta / \|\zeta_{i}\|)
  & \text{if} & \|\zeta_{i}\|>\gamma_{j}  \\
   0   & \text{if} &   \|\zeta_{i}\| \leq \gamma_{j}
\end{array}\right.
\quad , \quad  i=1\dots\nvg \ ,
\end{equation}
so that~:
\begin{equation}
\label{402-4}
 \|S_{j}(\zeta)\|^{2}  =  \displaystyle \sum_{i = 1}^{\nvg}  [\|\zeta_{i}\|-\gamma_{j}]_{+} ^{2} \ .
\end{equation}

\begin{Proposition} 
\label{pro resolution cl sgb}
The solution $(X^{*}Z^{*})$ of the GSMV formulation (\ref{350}) can be obtained in two steps~:
\begin{enumerate}
  \item Determine $X^{*}=[x^{*}_{1} \dots x^{*}_{\nl}]$ which maximizes over $\stns$ the convex function~:
\begin{equation}
\label{404}
   \Fmugamma(X)  = \sum_{j=1\dots\nl} \mu_{j}^{2}  \, \sum_{i=1}^{\nvg} \big[\|A_{i}^{T}x_{j}\|-\gamma_{j} \big]_{+}^{2} 
  = \sum_{j=1\dots\nl}\mu_{j}^{2} \, \|S_{j}(A^{T}x_{j})\|^{2}  \ . \hspace{2em}
\end{equation}
   \item Compute   $t_{j}^{*}=S_{j}(A^{T}x_{j}^{*})$ for $j=1\dots\nl$, and
   define $Z^{*}=[z^{*}_{1} \dots z^{*}_{\nl}]$ by~:
\begin{equation}
\label{412}
 z_{j}^{*}= 
 \left\{\begin{array}{cl}
    t^{*}_{j}/\|t^{*}_{j}\|   &   \mbox{ if } t^{*}_{j} \neq 0  \\
    0   & \mbox{ if } t^{*}_{j}=0  
\end{array}\right.  
\quad , \quad  j= 1 \dots \nl \ .
\end{equation}
When $\gamma_{j}\rightarrow 0$, one sees that $t_{j}^*\rightarrow A^{T}x_{j}^{*}$, so $t_{j}^{*}$ can be understood as a perturbation of $A^{T}x_{j}^{*}$ caused by the sparsity inducing parameter $\gamma_{j}$.
\end{enumerate}
 The condition~:
\begin{equation}
\label{414}
 \min_{j=1 \dots \nl}  \gamma_{j } < \max_{i=1\dots \nvg} \|A_{i}\|_{2}\ ,
 \end{equation}
on the regularization weights $\gamma_{j}$ ensures that at least one of the $t_{j}^{*}$ and $z^{*}_{j}$ are non zero, and hence that the value of the maximum in (\ref{404}) is strictly positive.  
\end{Proposition}
The proof of this proposition is given in Section \ref{proof of pro resolution cl sgb} of the Appendix.

\subsection{A  Block algorithm for the GSMV formulation}
\label{resolution of the group-sparse component/loading block formulation}

\subsubsection{Algorithm}
\label{num res algo}

The function $F(X)$ in proposition~\ref{pro resolution cl sgb} is convex and the Stiefel manifold $\stns$ compact, so one can use  
\cite[Algorithm 1 page 526]{JNRS2010},  which computes $X^{*}$ as the limit 
of the sequence $X_{k}$ defined by:
\begin{equation}
\label{414-1}
 X_{k+1} \mbox{ maximizes } \{ F(X_{k})+\langle \nabla F(X_{k}),Y-X_{k} \rangle \}
 \mbox{ over } {Y\in \stns} \ .
\end{equation}

\noindent The maximizer of $\langle \nabla F(X_{k}),Y \rangle$ over all $Y\in \stns$ in
\eqref{414-1} is the polar  of the $\ns \times \nl$ matrix  $\nabla F(X)= 2ATN^{2}$, where
\begin{equation}
\label{472}
 T= \big[ t_{1}  \dots t_{\nl}\big] \in \R^{\nvs\times \nl} \quad , \quad
 N=\diag(\mu_{1}, \dots,\mu_{\nl}) \in \R^{\nl\times \nl} \ .
\end{equation}
This gives the Group-Sparse Maximum Variance  block algorithm~:

\vspace{1.5ex}

\begin{minipage}{\textwidth}
\textbf{GSMV block algorithm}

\noindent \begin{tabular}{lcl}
\textbf{input} & : & $X_{0}\in \stns$ \\
\textbf{output} & : &  $X_{n}$ (approximate solution) \\

\textbf{begin} &  &  \\
   \hspace{1em}\begin{tabular}{| l}
    $0$  $ \longleftarrow$  $k$  \hspace{-25em}  \\
    \textbf{repeat} \\
       \hspace{1em}\begin{tabular}{| lcl}
       $T_{k}$ & $\longleftarrow$ & group-thresholding of $A^{T}X_{k}$ according to (\ref{402}) \hspace{-25em} \\
             $G_{k}$ & $\longleftarrow$ & $\nabla F(X_{k})=2AT_{k}N^{2}$ \hspace{-25em} \\
       $X_{k+1}$ & $\longleftarrow$ & $\polar(G_{k})$ \hspace{-25em} \\
        $k$ & $\longleftarrow$ &  $k+1$ \hspace{-25em}
       \end{tabular} \\
    \textbf{until} a stopping criterion is satisfied \hspace{-35em}
    \label{}
    
   \end{tabular}  \\
   
\textbf{end}   &  &
\end{tabular}
\end{minipage}
Once $X^{*}$ has been determined, $Z^{*}$ is computed from  \eqref{412} by applying the group-thresholding operator \eqref{402} to $A^{T}X^{*}$.

One iteration requires the computation of  the matrix products  $A^{T}X$ and $AT$, the group thresholding of the $p\times \nl$ matrix  $A^{T}X$, and the polar decomposition of  the $\ns \times \nl$ matrix $ATN^{2}$

By construction, the {GSMV block algorithm} has the convergence properties of 
\cite[Algorithm 1 page 526]{JNRS2010}~: it limit points are all stationary points of $F$, but non-necessarily local maxima, which is the best one can expect from local gradient methods.

\subsubsection{Initialization}
\label{algo initialization}

The function $F$ has many local maxima, so in order to limit the odds that the algorithm converges to a local maximum and/or produces loadings in the wrong order, we have chosen in all numerical experiments - deflation as well as block algorithms - to perform first an unconstrained PCA by running the algorithm with $\gamma_{j}=0$ for all $j$. Then we can use  the left singular vectors $[\vv_{1} \dots \vv_{\nl}]$ as initial value~$X_{0}$, and the singular values to balance the regularization level between the loadings - c.f. section~\ref{choice reg par}.

\subsubsection{Choice of regularization parameters}
\label{choice reg par}

In order to impose a similar level of regularization on all loadings, we have chosen to 
fit each {\em sparsity parameter} $\gamma_{j}$ to  the norm of the vector $A^{T}x_{j}$ it is in charge of thresholding. This norm is simply estimated by 
its initial value $\|A^{T}x_{j}^{0}\| = \|A^{T}\vv_{j}\| = \|\sigma_{j}\ww_{j}\| = \sigma_{j}$. To this effect we define for each component a {\em nominal sparsity parameters} $\gamma_{j,max}$  by~:
\begin{equation}
\label{nr4}
  \gamma_{j,max}= \frac{\sigma_{j}}{\sigma_{1}} \, \gamma_{max}
 \quad \mbox{where} \quad   \gamma_{max} \egaldef  \max_{i=1\dots \nvg} \|A_{i}\|_{2}
 \quad \mbox{as defined in (\ref{414})} \ ,
\end{equation}
and a {\em reduced sparsity parameters} $\lambda_{j}$ by~:
\begin{equation}
\label{nr4a}
   \lambda_{j} = \gamma_{j} / \gamma_{j,max} \quad , \quad j=1\dots \nl \ .
\end{equation}
Moreover, in the usual situation where no  a priori information on the sparsity of the underlying loadings is known, one can apply the same level of regularization for all components simply by using 
 the same reduced parameters 
$\lambda \in [0,1]$ for all loadings~:
\begin{equation}
\label{nr5}
  \displaystyle  0 \leq  \lambda = \lambda_{1}=\dots= \lambda_{\nl} \leq 1 \ .
\end{equation}
When this choice is done, the regularization parameters satisfy 
\begin{equation}
\label{ }
 0 \leq \gamma_{j} <  \gamma_{j,max} \leq \gamma_{max} 
  \quad , \quad j=1\dots \nl \ ,
\end{equation}
 which of course satisfies condition (\ref{414}), and up to $\nl-1$ loadings may vanish at the optimum.

\subsubsection{Choice of weights $\mu_{j}$}
\label{choice weight muj}

According to Proposition \ref{pro 1-Z}, we have considered two options~: 

\vspace{2ex}

\noindent $\bullet$ \diff: here we use strictly decreasing weights $\mu_{j}$, for example~:
\begin{equation}
\label{nr7}
 \mu_{j}= 1/j  \quad j=1\dots \nl \ ,
\end{equation}
in order to relieve the underdetermination which happens for equal $\mu_{j}$ at 
$\lambda=0$ and to drive the optimization, when $\lambda>0$, towards a minimizer $X^{*}$ which is ``close'' to the $\nl$ first left eigenvectors $[\vv_{1},\dots,\vv_{\nl}]$

\noindent $\bullet$ \same: but we have also tested  the behavior of the algorithm for equal weights~:
\begin{equation}
\label{nr6}
 \mu_{j}=1  \quad j=1\dots \nl \ ,
\end{equation}
as it  corresponds exactly to the maximization of  a sparse penalized version of the variance 
$\var(AZ)$ of Definition \ref{def 1}.

\subsection{A deflation  group-sparse algorithm for PCA}
\label{a group-sparse deflation algorithm}

For comparison purpose with the above block algorithm, we recall here a group-sparse deflation algorithm~:
\begin{eqnarray}
\mbox{Set } A_{0} & = & A \ ,\ z_{0}=0 \ , \mbox{ and compute, for $j=1 \dots m$ :} \label{476} \\
A_{j} & = &  A_{j-1} (I_{\nvs}- z_{j-1}z_{j-1}^{T}) \label{477} \\
z_{j} & = & \arg \!\max_{\|z\|=1}(\|A_{j}z\| - \gamma_{j}\|z\|_{1}) \ \label{478}
\end{eqnarray}
The optimization problem (\ref{478}) coincides with the GSMV block formulation (\ref{350})  written for $\nl=1$. Hence we shall implement the deflation  algorithm by applying the {GSMV block algorithm} of Section \ref{num res algo} with $\nl=1$ (one single sparse loading) iteratively to each deflated matrix $A_{j}$.

\subsection{Specification to sparse PCA of a mixture of numerical and categorical variables}
\label{specification for mixed variables}

The columns of $A$ are now are either centered (or standardized) numerical variables or centered binary variables coding the levels of a categorical variable. Let $p_1$ denote the number of numerical variables and $p_2$ the number of categorical variables.  Each categorical variable $j$ has $q_j$ levels and $q=\sum_{j=1}^{p_2} q_j$ is the total number of levels. The number of columns of $A$ is then $p=p_1+q$ and the dimension of the matrix $A$ is  then $n \times (p_1+q)$. 

Let $\| . \|_{M,N}$ denote the generalized Frobenius or Hilbert-Schmitt norm
 on the space of $\ns \times \nvs$ matrices~:
\begin{equation}
  \|A\|_{M,N}^{2} 
  =\tr(A^TNAM)
  =\sum_{j=1\dots r}\sigma_{j}^{2}\ ,
\end{equation}
where $M$ and $N$ are symmetric positive definite matrices and the $\sigma_{j}$'s are the singular values of the  \emph{generalized singular value decomposition} (GSVD) of $A$~:
\begin{equation}
\begin{array}{l}
    A=\V\Sigma \W^{T}\quad \mbox{ with } \quad \V^{T}N\V=I_{r}\quad,\quad 
    \W^{T}M\W=I_{r} \ , \\
 \Sigma=\diag(\sigma_{1}, \dots,\sigma_{r}) = \mbox{$r \times r$ matrix with } \sigma_{1}\geq \sigma_{2}\geq \dots \geq \sigma_{r} >0 \ .
 \end{array}
\end{equation}
The columns $\vv_{1} \dots \vv_{r}$ of $\V$ and  $\ww_{1} \dots \ww_{r}$ of $\W$ are the left and right \emph{generalized singular vectors}. 

PCA of a mixture of numerical and categorical data (PCAmix)  finds a number $\nl \leq r$ of combinations $z_{j},j=1,\dots \nl$ of the $\nvs=p_1+q$ columns of $A$  such that the variables $y_{j}=AMz_{j}, j=1 \dots \nl$ are uncorrelated and explain an as large as possible fraction of the variance $\|A\|_{M,N}^{2}$ of the data. Here, $N$ is the diagonal matrix of the weights of the $n$ observations ($N=\mathbb{I}_n$ when all observations are weighted by 1) and $M$ is the diagonal matrix of the weights of the $p_1+q$ columns of $A$:
$$M=\mbox{diag}(1,\dots,1,\frac{n}{n_1},\dots,\frac{n}{n_q})$$ 
The $p_1$ first columns (numerical variables) are weighted by 1. The $q$ last columns (levels) are weighted by the inverse of the frequencies $\frac{n_1}{n} \ldots \frac{n_q}{n}$ of the $q$ levels. This metric $M$ (between observations) is the standard Euclidean distance for numerical variables and a weighted Euclidean distance (in the spirit of the $\chi^2$ distance) for the levels (to give more importance to rare levels). The loadings and components solutions to the PCAmix  problem are then given by~:
 \begin{equation}
Z = V_{\nl} \quad ,  \quad Y= AMV_{\nl}
\end{equation}
where $V_{\nl}$ contains the $\nl$ first right  \emph{generalized singular vectors}.

Sparse PCA of a  mixture of numerical and categorical variables (sparsePCAmix) is then performed via group-sparse PCA as follows. Groups of variables are specified according to the nature of the variables: each numerical variable defines a group of size 1 and each categorical variable with  $q_j$ levels  defines a group of size $q_j$. Group-sparse PCA is then applied to the matrix ${\tilde A}=N^{1/2} A M^{1/2}$ to find a solution ${\tilde Z}$ and this solution is transformed back to the original scale to find the sparse loadings $Z=M^{-1/2}{\tilde Z}$. In this way  the loadings of the levels of a categorical variable are  simultaneously set to 0.

\section{Numerical results}
\label{numerical results}

The GSMV block algorithm of Section \ref{resolution of the group-sparse component/loading block formulation} and its adaptation to mixed data of Section  \ref{specification for mixed variables}, as well as the deflation algorithm of Section \ref{a group-sparse deflation algorithm}, have been implemented in the R~packages
``sparsePCA'' and ``sparsePCAmix'' available at \url{https://github.com/chavent/}.
All numerical results of this section have been produced using these codes.

\subsection{Performance indicators}
\label{performance indicators}

The first measure of performance is the variance $\var Y$ explained
by the (usually non orthogonal) $\nl$ components $y_{j}$. We use here for $\var Y$ the \emph{optimal projected variance} as suggested in Definition \ref{def 1}, which we recall here~:
\begin{equation}
\label{370aa}
\var Y =  \max_{X\in\stns} \sum_{j=1\dots\nl} \langle y_{j}\,,x_{j}\rangle^{2} \ .
\end{equation}
As shown in Section \ref{how to define the explained variance} above, this variance is optimal in the sense that it produces the largest explained variance among a large class of possible definitions.

We display  the dimensionless 
\emph{proportion of explained variance} (pev) \eqref{nr25}, which measures the loss in explained variance induced by sparsity compared to PCA, as well as the individual contributions $pev_{j}$ of each components~:
\begin{equation}
\label{536d}
pev =  \sum_{j=1}^{\nl} pev_{j}  = \var Y / \varpca \quad , \quad
pev_{j}=\|y_{j}^{\prime}\|^{2} / \varpca \quad j=1\dots \nl \ ,
\end{equation}
where the $y_{j}^{\prime}$ are the ``modified loadings'' \eqref{532}  which enter the definition of projected variances.

The  {\em orthogonality  of components} will be measured by 
the $\nl$-dimensional volume $\mathrm{Vol}(Y)=|\det(P)|$ of the 
parallelepiped constructed on the columns of $Y$,
where $Y=UP$ is  the polar decomposition of $Y$. One defines then a  {\em dimensionless orthogonality measure} 
 $\mathrm{volume}$ of $Y$ by~: 
\begin{equation}
\label{nr20} 
 0 \leq  \mathrm{volume}  = \mathrm{Vol}(Y) \big /  \prod_{j=1 ... \nl}   \|y_{j}\| \leq 1
 \quad \text{ where } \mathrm{Vol}(Y)=|\det(P)| \ .
\end{equation}
In the case where one $y_{j}$ or more vanish, which may happen as up to $\nl-1$ optimal $z_{j}$ may be equal to zero (c.f. Section \ref{choice reg par}), the volume index is computed on the non-zero remaining $y_{j}$s.

When the algorithm is run on simulated data, where the true sparsity pattern is known,
the {\em adequation of the sparsity structure} of the estimated loadings $Z$ to that of the underlying $Z_{true}$ is measured by~:
\begin{itemize}
  \item[-] the {\em true positive rate} (tpr)~: proportion of zero entries of $Z_{true}$ retrieved as $0$ in $Z$,
  \item[-] the  {\em false positive rate} (fpr)~: proportion of non zero entries of $Z_{true}$ retrieved as $0$ in~$Z$.
\end{itemize}
These quantities can be evaluated loading by loading (i.e. on the columns 
of $Z$), or globally over all loadings (i.e. on the whole matrix $Z$).

The {\em subspace proximity between $Z_{true}$ and $Z$} will be measured by  
 $\rv(Z,Z_{true})$ where $\rv(X,Y)$ is the $\rv$-index defined following \cite{escoufier1973traitement} and \cite{abdi2007rv} by~:
\begin{equation}
\label{nr10}
 \rv(X,Y)= \frac{\|X^{T}Y\|_{F}^{2}}{\| X^{T}X \|_{F} \|Y^{T}Y\|_{F}} 
 	     = \frac{ \langle X^{T}X , Y^{T}Y \rangle_{F} } {\| X^{T}X \|_{F} \|Y^{T}Y\|_{F}} 
 	     = \rv(Y,X) \ .
\end{equation}
The first formula is used to compute $\rv$, and  the second implies that~:
\begin{equation}
\label{nr15}
   0 \leq  \rv(X,Y)  \leq 1 \ .
\end{equation}

\subsection{Simulated data}
\label{num res simulated data}

\subsubsection{Data generation}
\label{dta generation}

In order to test the ability of the algorithms  to retrieve group-sparse singular vectors, we have generated \emph{two sets of synthetic data matrices $A$} with $\ns=300$ samples and  $\nvs=20$ variables each, which share the four first sparse underlying right singular vectors given Table~\ref{fig 3} (the ``underlying loadings $Z_{true}$''). There are hence  $\nvg=5$ group variables made of $\dv_{j}=4$ scalar variables each. The two sets differ by the underlying eigenvalues~:
\begin{itemize}
  \item[-]  the ``close eigenvalue'' set is associated to eigenvalues $200,180,150,130, 1 ... 1$,
  \item[-] the ``different eigenvalue'' set is associated to eigenvalues $200,100,50,20, 1 ... 1$.
\end{itemize}
We have simulated for each case 100  matrices $A$.
More precisely, we have followed the procedure proposed by \cite{SH2008} and \cite{JNRS2010} to generate data matrices $A$ by drawing $\ns$ samples from a zero-mean distribution with covariance matrix $C$ defined  by  $C=\W_{true}\Sigma_{true}^{2}\W_{true}^T$~, where $\Sigma_{true}^{2}$ is the diagonal matrix of the chosen eigenvalues, and $\W_{true}$ is the $\nvs \times \nvs$ orthogonal  matrix defined by the QR-decomposition $[Z_{true},U]=\W_{true}R$, where $U$ of dimension $\nvs \times (\nvs-\nl)$ is randomly drawn from $U(0,1)$. Notice that, by definition of the QR-decomposition, the $\nl$ first columns of $\W_{true}$ coincide with $Z_{true}$.
 
Additional experiments with 3000 samples instead of 300 have been performed in the (more difficult) case of the ``close eigenvalue'' set of synthetic data matrices. 
\begin{table}[ht]
\centering
\scriptsize
\begin{tabular}{rrrr}
  \hline
0.253 & 0.000 & 0.000 & 0.220 \\ 
-0.253 & 0.000 & 0.000 & 0.220 \\ 
 0.253 & 0.000 & 0.000 & 0.220 \\ 
  -0.253 & 0.000 & 0.000 & 0.220 \\ 
  0.000 & 0.393 & 0.416 & 0.000 \\ 
0.000 & 0.393 & 0.416 & 0.000 \\ 
0.000 & -0.393 & 0.416 & 0.000 \\ 
 0.000 & -0.393 & 0.416 & 0.000 \\ 
-0.211 & 0.262 & 0.000 & 0.183 \\ 
-0.211 & 0.262 & 0.000 & -0.183 \\ 
0.211 & 0.262 & 0.000 & 0.183 \\ 
 0.211 & 0.262 & 0.000 & -0.183 \\ 
0.168 & 0.000 & 0.000 & -0.367 \\ 
0.168 & 0.000 & 0.000 & -0.367 \\ 
0.168 & 0.000 & 0.000 & -0.367 \\ 
0.168 & 0.000 & 0.000 & -0.367 \\ 
0.337 & 0.164 & 0.277 & 0.183 \\ 
0.337 & 0.164 & -0.277 & 0.183 \\ 
0.337 & -0.164 & 0.277 & 0.183 \\ 
0.337 & -0.164 & -0.277 & 0.183 \\ 
   \hline
\end{tabular}
\captionof{table}{The underlying  $\nvs \times \nl$  group-sparse block of loadings $Z_{true}$}
\label{fig 3}
\end{table}

\subsubsection{Block versus deflation}
\label{block vs deflation}

We compare here the performances of three group-sparse algorithms~: 
\begin{itemize}
  \item \defl: the deflation algorithm described in Section \ref{a group-sparse deflation algorithm}, 
  \item \diff: the GSMV block algorithm of Section \ref{resolution of the group-sparse component/loading block formulation} with $\mu_{j}=1/j$
  \item \same:
as above, but with $\mu_{j}=1$ for all $j$.
\end{itemize}
The same reduced sparsity parameter $\lambda$ is chosen for all loadings as proposed in section~\ref{choice reg par}, and 
its influence  is explored by letting it vary from $0$ to $1$ by steps of 0.01. 
The performances of the three algorithms are first compared in Figures  \ref{fig 6}, \ref{fig 11}, \ref{fig 11a} and \ref{fig 10} on the ``different eigenvalues'' set of data matrices $A$ with 300 samples.

Figures \ref{fig 6} shows the mean values of  the $\rv$-index (proximity to the underlying $Z_{true}$) (left), the percentage of explained variance pev defined by  \eqref{370aa} \eqref{536d} (center) and the volume measure of orthogonality of components (right) as a function of $\lambda$. One sees that \diff  performs better than \defl 
for the three investigated indexes over essentially the whole range of $\lambda$. 
The most significative gain occurs for the $\rv$-index, where a gain of $10\%$ can be observed for values of $\lambda$ as small as $0.4$.
Unsurprisingly, the \same algorithm, where the objective function is a sparsified version  of the variance $\var Y$ itself,  produces a significantly larger $pev$, at the expense of a worse proximity $\rv$-index to the underlying $Z_{true}$ and of less orthogonal loadings. Notice the somewhat chaotic behavior of the volume orthogonality index (Figure \ref{fig 6} right, green dotted line) corresponds to the vanishing of the first loading for $\lambda \simeq 0.3$ (Figure \ref{fig 11a} right, black dotted line). 

The boxplots of  Figure \ref{fig 11} for $\lambda=0.2$ confirm that the results are slightly better with \diff than with  \defl with similar dispersions.

Figure \ref{fig 11a} (left and center) shows that \defl and  \diff algorithms produce the right number of zeroes for each component for a range of $\lambda$ between $0.1$ and $0.3$, in opposition to the \same algorithm (right), which adds quickly too many zeros to the first loading.

\noindent
\begin{minipage}[c]{\textwidth}
\centering
\resizebox{40em}{!}{\includegraphics{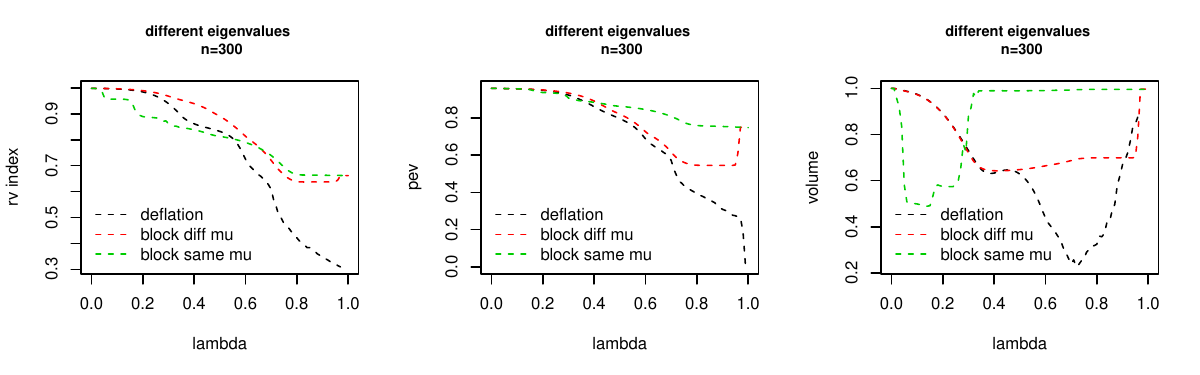}}
\captionof{figure}{Mean value of the proximity index rv to the underlying  loadings $Z_{true}$ (left),
proportion of explained variance defined by \eqref{370aa} \eqref{536d} (center) and orthogonality measure volume (right) as function of $\lambda$,
for the three algorithms.}
\label{fig 6}
\end{minipage}

\noindent
\begin{minipage}[c]{\textwidth}
\centering
\resizebox{40em}{!}{\includegraphics{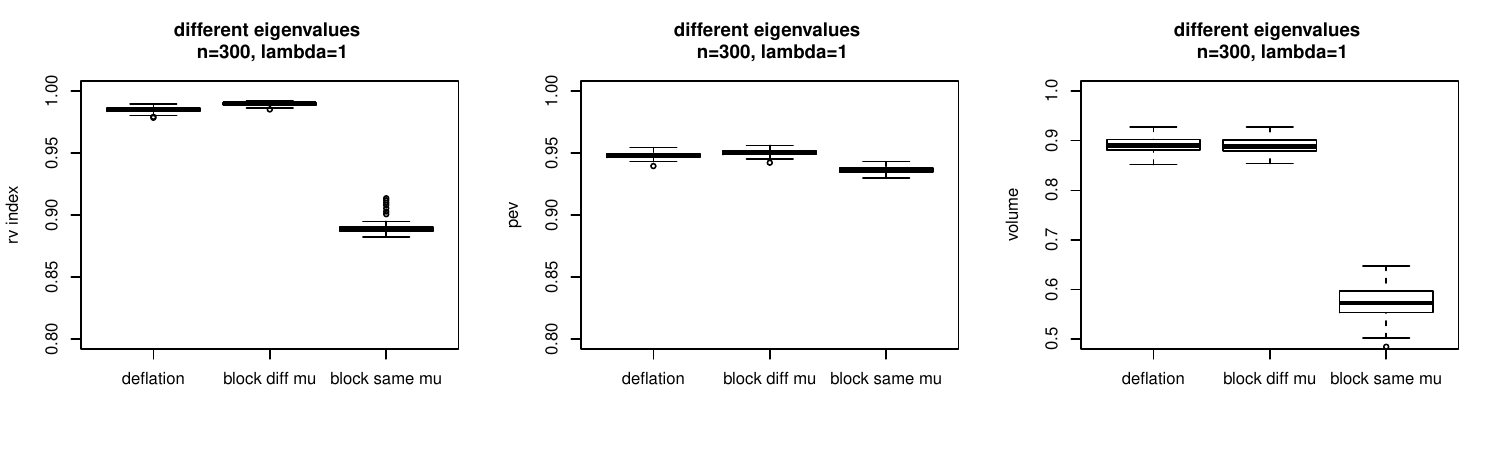}}
\captionof{figure}{Boxplots for the proximity index rv to the underlying  loadings $Z_{true}$ (left), the proportion of explained variance defined by \eqref{370aa} \eqref{536d} (center) and the orthogonality measure volume (right) for $\lambda=0.2$, for the three algorithms.}
\label{fig 11}
\end{minipage}

\noindent
\begin{minipage}[c]{\textwidth}
\centering
\resizebox{40em}{!}{\includegraphics{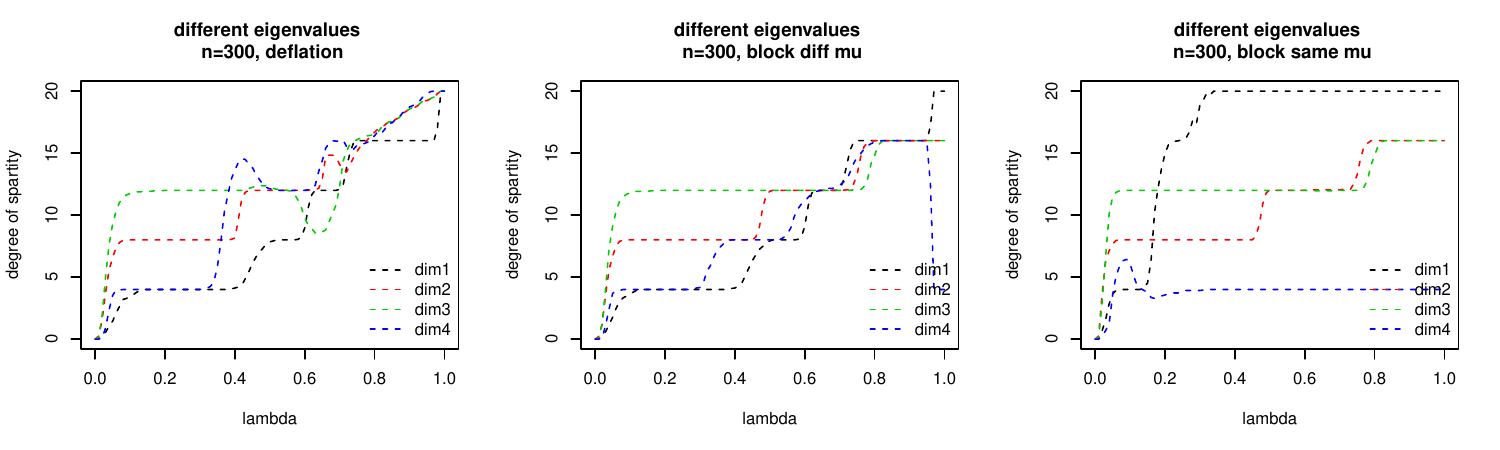}}
\captionof{figure}{Mean value of the number of zeroes in each loading as function of $\lambda$, for the three algorithms.}
\label{fig 11a}
\end{minipage}


As a check for the choice (\ref{nr4}) (\ref{nr4a}) (\ref{nr5}) of the sparsity parameters 
$\gamma_{j}$, we have plotted in Figure \ref{fig 10} the decay, as a function of $\lambda$, of the contributions $\mathrm{pev}_{i}$ of each sparse component to the explained variance $\var Y$, as defined by \eqref{370aa} \eqref{536d}. As one can see, the decrease is roughly similar, which indicates that the relative size of the 
$\gamma_{j}$ is correctly chosen.

\noindent
\begin{minipage}[c]{\textwidth}
\centering
\resizebox{40em}{!}{\includegraphics{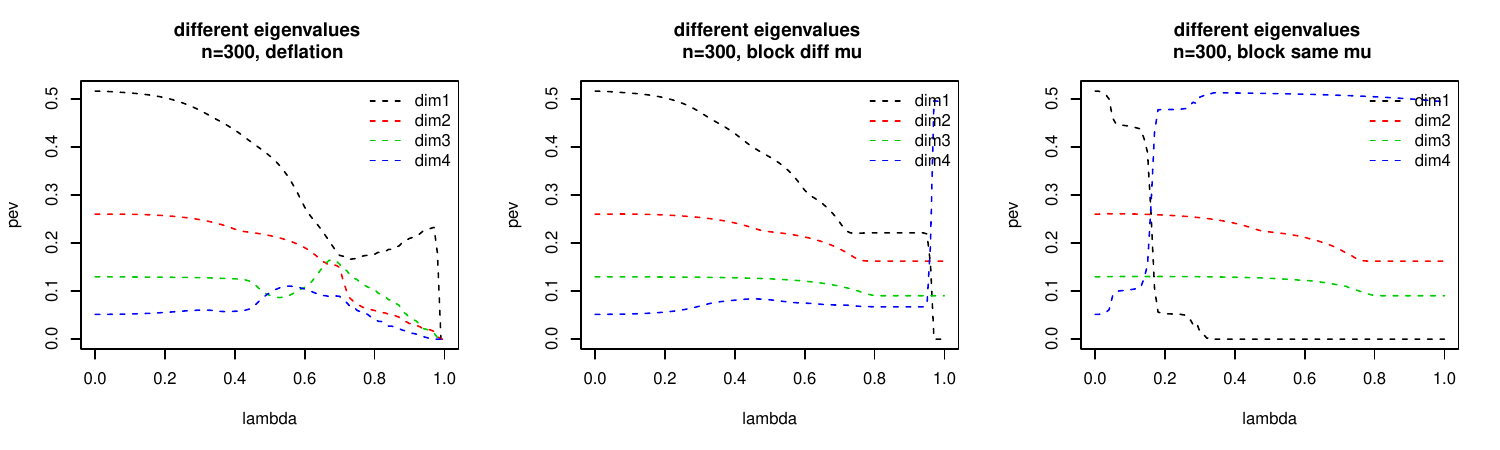}}
\captionof{figure}{Contribution of each component to the explained variance $\var Y$ as function of $\lambda$ for the three algorithms in the case of 300 samples.}
\label{fig 10}
\end{minipage}

\noindent
\begin{minipage}[c]{\textwidth}
\centering
\resizebox{40em}{!}{\includegraphics{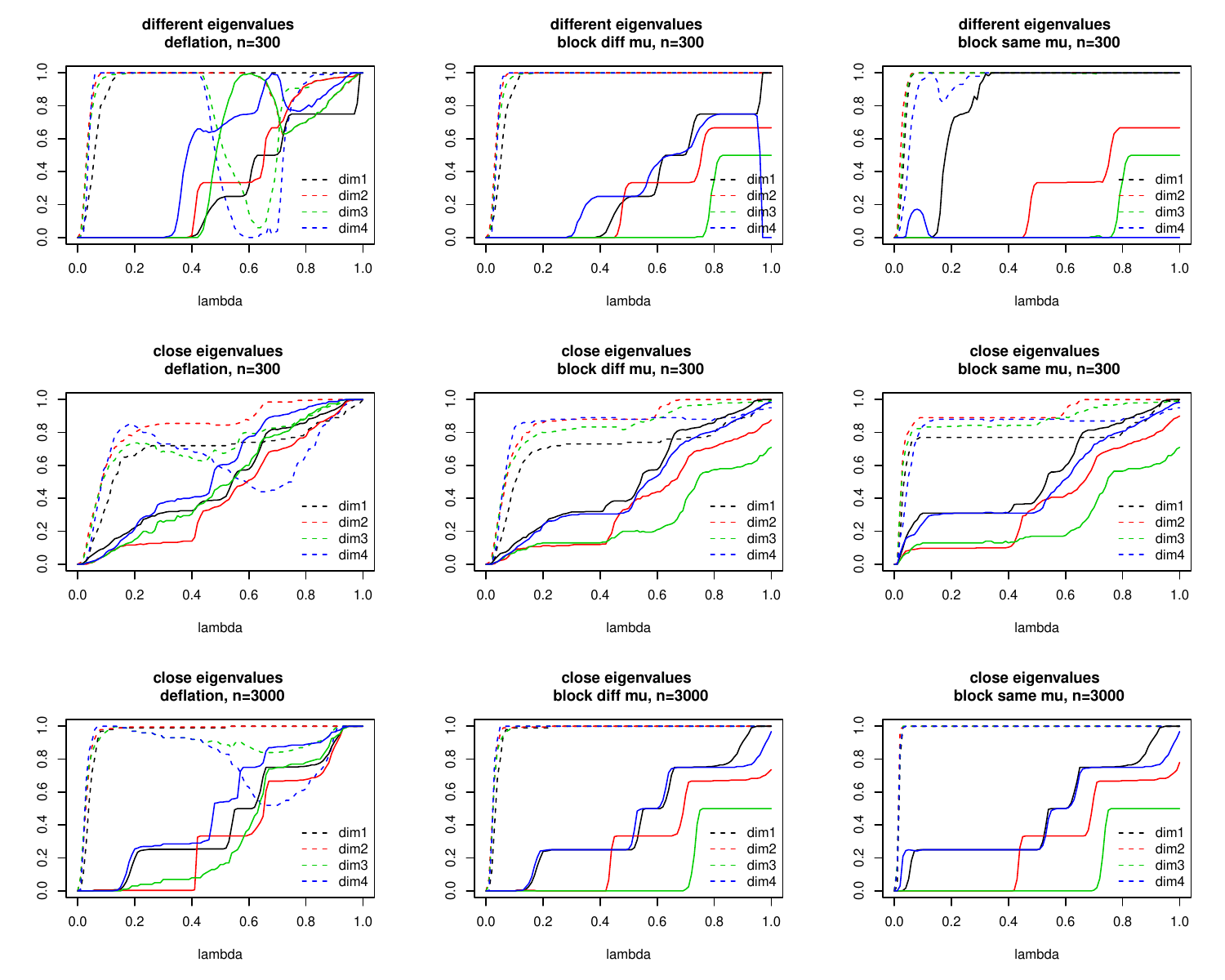}}
\captionof{figure}{Mean true positive rates (dotted lines) and false positive rates (full lines) for each sparse loading versus reduced sparsity parameter $\lambda$. From left to right: \defl, \diff,  
\same. Top~: different eigenvalues and $n=300$, center~: close eigenvalues and  $n=300$, bottom~: close eigenvalues and $n=3000$.}
\label{fig 4}
\end{minipage}

In order to get a more detailed view of the  ability of the algorithms to retrieve the exact sparse structure of $Z_{true}$,
we show in Figure \ref{fig 4} the mean values of $tpr$ (true positive rate) and $fpr$ (false positive rate) for loadings obtained by applying  the three algorithms (from left to right)  to  two sets of 300 samples matrices   $A$  
(top~: different eigenvalues, center~: close eigenvalues) and one set of 3000 samples matrices $A$ (bottom~: close eigenvalues). 
With this representation the sparsity pattern is perfectly recovered for the values of $\lambda$ such that $tpr=1$ and $fpr=0$.
As expected, increasing $\lambda$ increases the global true positives, at the expense of more false positive. 

The top row of the figure shows that in the case of ``different eigenvalues'', both the \defl and \diff algorithms are able to retrieve the exact group sparse structure of $Z_{true}$ for a large interval $0.1 \leq \lambda \leq 0.3$, as expected from Figure \ref{fig 11a}. 
The center row shows that the ``close eigenvalues`` case is more difficult, as the exact sparsity structure is never exactly recovered by any of the three methods.
The bottom row shows that the difficulty related to ``close eigenvalues''  can be overcome by increasing the number of samples from 300 to 3000~: both the \defl  and the \diff algorithms are now able to retrieve, even in the mean, the exact sparsity structure of $Z_{true}$ for $\lambda \simeq 0.1$. 
The \diff algorithm shows hence a slight advantage over the deflation in that the
 tpr grow slower and fpr grow faster for small values of $\lambda$, and that the tpr are less erratic for large large values of $\lambda$.
And in all theses case,  the \same algorithm performs the worst with its tendency to add too quickly wrong zeroes.

We conclude this section by comparing computation time and iteration numbers. We show  in Figure~\ref{fig_comptime} their mean values over the  $100$ random matrices $A$ for  the more difficult ``close eigenvalues'' case with 300 samples for 
$\lambda=0.2$. One sees that \diff  and \same are three times faster than \defl$\!\!\!$, and that \defl performs more iterations, which is expected as it repeats the iterations at each deflation step. Computation times are multiplied by $10$ for the same case with $3000$ samples; iteration numbers are down to $3\times 4=12$ for \defl and $3$ for \diff and \same.

\noindent
\begin{minipage}[c]{\textwidth}
\centering
\resizebox{40em}{!}{\includegraphics{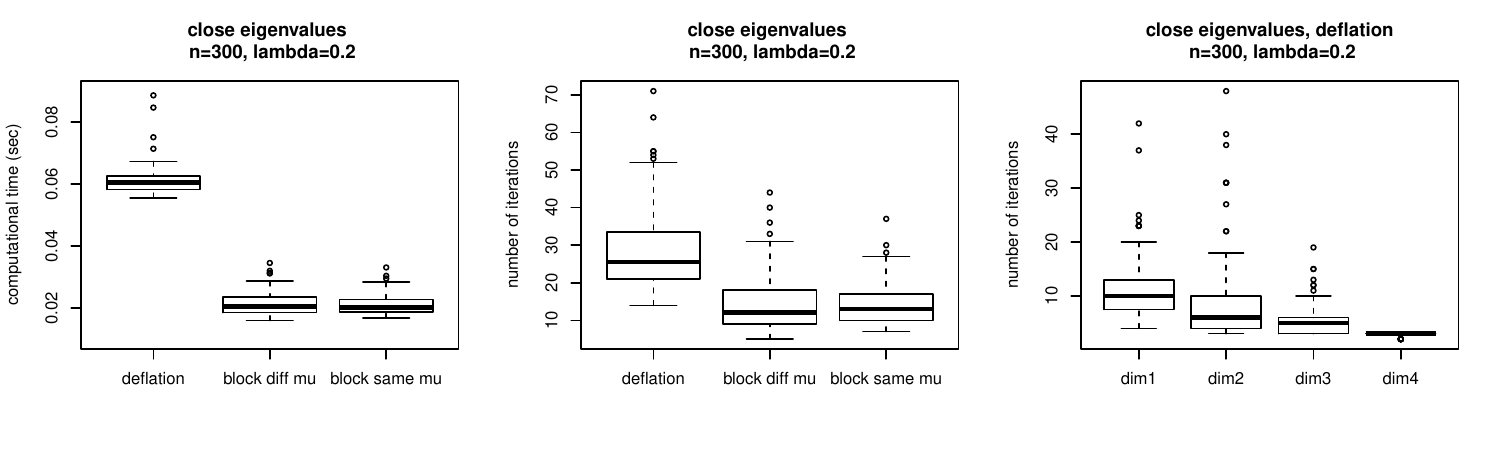}}
\captionof{figure}{Computational time and iteration numbers for the ``close eigenvalues'' case with 300 samples for the three methods and $\lambda = 0.2$.}
\label{fig_comptime}
\end{minipage}

\vfill
$\ $

\subsubsection{Sparse versus group-sparse}
\label{sparse versus group-sparse}

We illustrate in Figure \ref{fig 7} the effect of imposing sparsity on group of variables rather than on single variables for the close eigenvalues case with $n=300$.
The top row shows the true positive rates and false positive rates for the three algorithms when no group sparsity information is available, to be compared to the bottom row, where group sparsity is taken into account, and which exhibits higher true positive rates and lower false positive rates.
It is hence important to use the group structure information, when available, as it helps greatly the algorithm to retrieve the sparsity structure of the underlying loadings.

\begin{minipage}[c]{\textwidth}
\centering
\resizebox{40em}{!}{\includegraphics{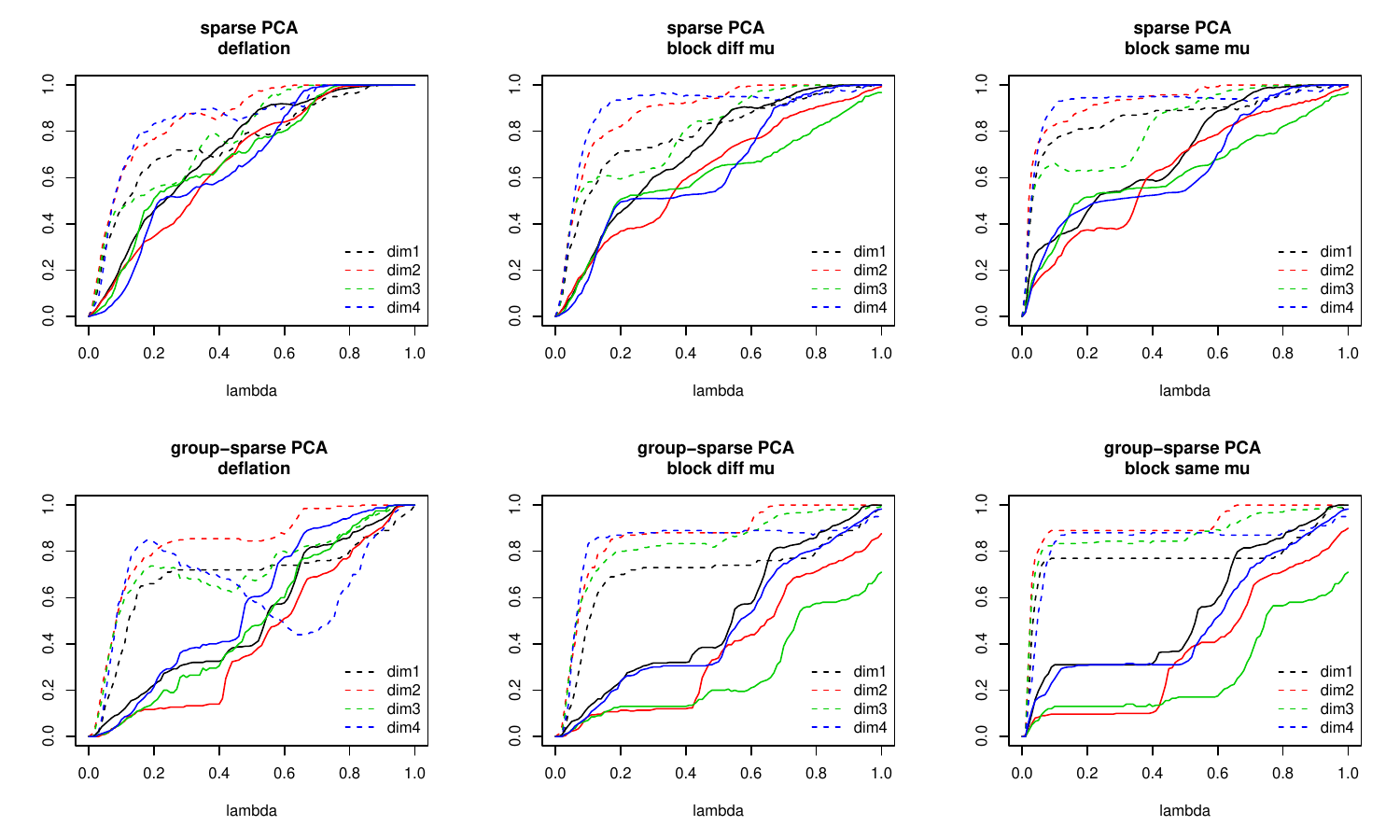}}
\captionof{figure}{True positive rates (dotted lines) and false positive rates (full lines) for each sparse loading versus reduced sparsity parameter $\lambda$ in the close eigenvalues case and $n=300$. Top: scalar variables, bottom: group variables.}
\label{fig 7}
\end{minipage}

\subsection{Real mixed data}
\label{numerical results real mixed data}

The heart disease dataset\footnote{https://archive.ics.uci.edu/ml/datasets/Statlog+Heart} \cite{Dua:2019} describes $n=270$ observations on a mixture of $p_1=6$ numerical variables and $p_2=7$ categorical variables. Each categorical variable has 2, 3 or 4 levels for a total of $q=19$ levels. In order to perform sparse PCA with this mixture of numerical and categorical variables (sparsePCAmix),  the matrix $A$ of size $270 \times 25$  ($25=p_1+q$) is build and transformed using specific metrics as described Section \ref{specification for mixed variables}. Then the group-sparse PCA algorithm \diff of Section \ref{resolution of the group-sparse component/loading block formulation} is applied with 
$\mu_{j}=1/j$ to the matrix $A$ with $g=13$ groups (one group for each variable): 
\begin{itemize}
\item[-] 6 groups of size 1 (for the 6 numerical variables),
\item[-] 7 groups of size 2, 3 or 4 (for the 7 categorical variables) where the size of the group corresponds to the number of levels.
\end{itemize}
The methods PCAmix (PCA of $A$) and sparsePCAmix (group-sparse PCA of $A$) are both implemented in the R package ``PCAmixdata'' available at \url{https://github.com/chavent/PCAmixdata}.

Table \ref{tab:varpcamix} gives the variance explained by the 3 first principal components  of PCAmix  (combinations of all columns of $A$) and shows that 35.41 \% of the variance of the data is explained by $m=3$ components.

\begin{table}[ht]
\centering
\begin{tabular}{rrrr}
  \hline
 & Eigenvalue & Proportion & Cumulative \\ 
  \hline
dim 1 & 3.22 & 17.87 & 17.87 \\ 
  dim 2 & 1.67 & 9.28 & 27.15 \\ 
  dim 3 & 1.49 & 8.26 & 35.41 \\ 
   \hline
\end{tabular}
\captionof{table}{Eigenvalues and proportion of variance explained by the 3 first principal components of PCAmix.}
\label{tab:varpcamix}
\end{table}

\begin{minipage}[c]{\textwidth}
\centering

\resizebox{40em}{!}{\includegraphics{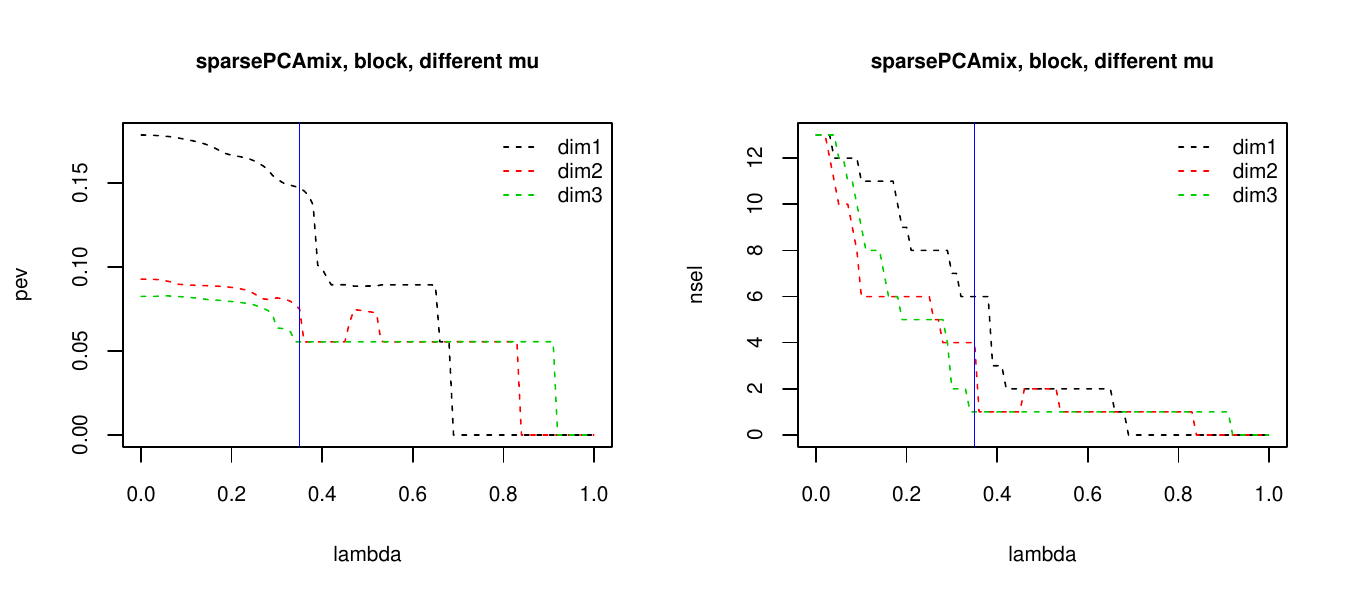}}
\captionof{figure}{Proportion of explained variance defined by \eqref{370aa} \eqref{536d} (left) and number of selected variables (right) as function of $\lambda$ for the sparsePCAmix method.}
\label{fig:evo_heart}
\end{minipage}

\vspace{2ex}

Figure \ref{fig:evo_heart} gives the proportion of explained variance (left) and of the number of selected variables (right) as a function of the reduced sparsity parameter $\lambda$ for sparsePCAmix with $m=3$ components.  This figure suggests to choose  $\lambda=0.35$. With this sparsity parameter, the explained variance over the 3 dimensions decreases from 35.41 \% to 27.76 \% (i.e. a lost of nearly 7\%). Morover 6 variables (2 numerical and 3 categorical) are selected to build the first component, 4 variables (3 numerical and 1 categorical)  are selected to build the second and the third one is build with a single categorical variable (see Table \ref{tab:loadings_heart}).

\begin{table}[ht]
\centering
\begin{tabular}{rrrr}
  \hline
 & dim 1 & dim 2 & dim 3 \\ 
  \hline
age & 0.00 & 0.40 & 0.00 \\ \hline
  blood\_pressure & 0.00 & 0.16 & 0.00 \\ \hline
  serum\_cholestoral & 0.00 & 0.86 & 0.00 \\ \hline
  max\_heart\_rate & 0.43 & 0.00 & 0.00 \\ \hline
  oldpeak & -0.51 & 0.00 & 0.00 \\ \hline
  number\_vessels & 0.00 & 0.00 & 0.00 \\ \hline
  sex=0 & 0.00 & 0.13 & 0.00 \\ 
  sex=1 & 0.00 & -0.13 & 0.00 \\ \hline
  chest\_pain=1 & -0.00 & 0.00 & 0.00 \\ 
  chest\_pain=2 & 0.08 & 0.00 & 0.00 \\ 
  chest\_pain=3 & 0.06 & 0.00 & 0.00 \\ 
  chest\_pain=4 & -0.14 & 0.00 & 0.00 \\\hline 
  fasting\_blood\_sugar=0 & 0.00 & 0.00 & 0.00 \\ 
  fasting\_blood\_sugar=1 & 0.00 & 0.00 & 0.00 \\ \hline
  resting\_results=0 & 0.00 & 0.00 & 0.00 \\ 
  resting\_results=1 & 0.00 & 0.00 & 0.00 \\ 
  resting\_results=2 & 0.00 & 0.00 & 0.00 \\ \hline
  induced\_angina=0 & 0.15 & 0.00 & 0.00 \\ 
  induced\_angina=1 & -0.15 & 0.00 & 0.00 \\ \hline
  slope=1 & 0.27 & 0.00 & 0.08 \\ 
  slope=2 & -0.21 & 0.00 & -0.31 \\ 
  slope=3 & -0.05 & 0.00 & 0.23 \\ \hline
  thal=3 & 0.13 & 0.00 & 0.00 \\ 
  thal=6 & -0.02 & 0.00 & 0.00 \\ 
  thal=7 & -0.11 & 0.00 & 0.00 \\ 
   \hline
\end{tabular}
\captionof{table}{Matrix $Z$ of sparse loadings for $\lambda=0.35$.}
\label{tab:loadings_heart}
\end{table}
\begin{minipage}[c]{\textwidth}
\centering

\vspace{-3ex}

\resizebox{40em}{!}{\includegraphics{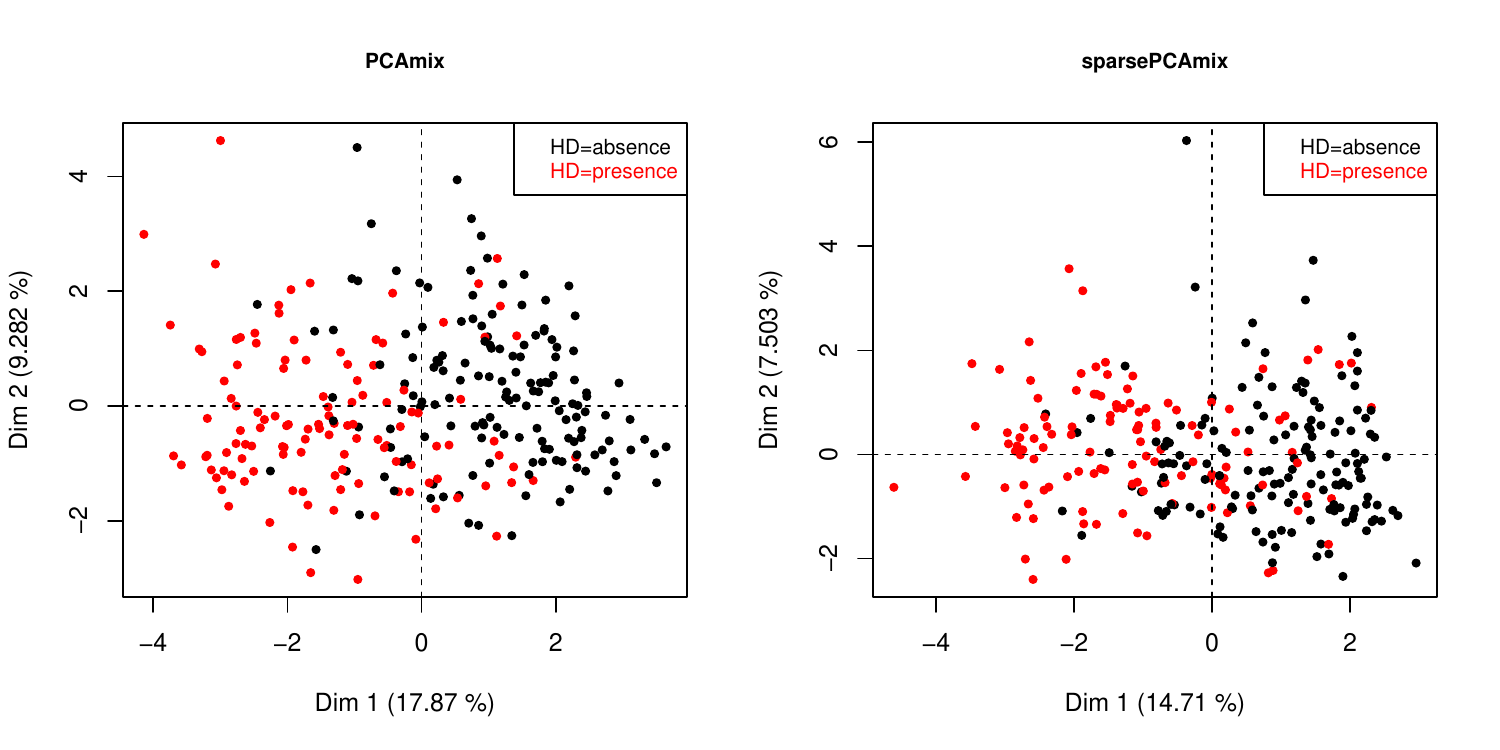}}
\captionof{figure}{PCAmix (left) versus sparsePCAmix for $\lambda=0.35$ (right).}
\label{fig:sparsePCAmix_heart}
\end{minipage}

Finally the $n=270$ observations are plotted Figure \ref{fig:sparsePCAmix_heart} according to the two first principal components of PCAmix (on the left) and of sparsePCAmix for $\lambda=0.35$ (on the right). Each observation is colored according to the binary variable of interest "Heart Diseases" (HD) (not used to build the components). The sparse first component on the right (build with 2 numerical and 4 categorical variables instead of 13 variables) keeps enough variance to discriminate between presence or absence of heart disease.

\section{Conclusion}
\label{conclusion}

We propose an approach to  \emph{Group-Sparse Block} Principal Component Analysis based on the maximization of the \emph{variance explained by non-necessarily orthogonal components}, with the objective of being able to analyze mixed data containing both numerical and categorical variables.

In a first step, we investigate the problem of defining the variance explained by non necessarily orthogonal components, and check existing and new definitions against their compatibility with the PCA  situation. As a result of this study, we propose to use the \emph{optimal projected variance}, which is larger than all projected variance, as the definition of  $\var Y$ for non orthogonal components.

Maximization of $\var(AZ)$ with respect to unit norm loadings $Z$ provides a new Block PCA formulation without orthogonality constraints on the loadings.
A Group-Sparse Maximum Variance (GSMV) block PCA formulation is naturally derived by an ad-hoc penalization by the  group-$\ell^{1}$ norm of the loadings.
We show that GSMV reduces to the  maximization a convex function over a Stiefel manifold, and  generalizes the $\ell^{1}$-algorithm of \cite{JNRS2010}. 

We propose a strategy for the choice of the regularization parametersin order to balance the sparsifying effort on the loadings, and show
numerical results on simulated data which confirm the expectation that GSMV with decreasing weights is more robust and performant than deflation for group-sparse PCA~: it produces steadily sparse loadings closer to the true underlying ones,
larger explained variance and better orthogonality of the components,
the tpr increase faster and the fpr slower, 
and  it ran approximately three times faster on our tests.

Numerical results show that the use of group sparsity allows a better retrieval of  
underlying sparse structures,
but the success is impacted by the heterogeneity  of the PCA eigenvalues~: problem with close eigenvalues are more difficult, but increasing the number of sample helps.
Application of  GSMV to the the interpretation of 
real Heart Disease data, has allowed to reduce the number of numerical and categorical variables required for the interpretation by a factor two.

\newpage

\section{Appendix }
\label{appendix}

\subsection{Generalized Rayleigh quotient}
\label{properties of generalized rayleigh quotient}
This is a classical result, see for example \cite{AMS2008} and 
\cite{Brockett1991}~:
\begin{Theorem}
\label{thm GRQ} 
Let the loadings $Z$ satisfy~:
\begin{equation}
\label{570-1}
  Z  =  [z_{1} \dots z_{\nl}]\in \R^{\nvs \times \nl}  \quad, \quad \rank \, Z=\nl \leq \rank A \egaldef r  \ .
\end{equation}
Then the generalized Rayleigh quotient
\begin{equation}
\label{570-0}
  \tr\{ (Z^{T}A^{T}AZ)(Z^{T}Z)^{-1}\}
\end{equation}
 satisfies~:
\begin{equation}
\label{570-2}
  \tr\{ (Z^{T}A^{T}AZ)(Z^{T}Z)^{-1} \} \leq \sigma_{1}^{2}+\dots+\sigma_{\nl}^{2}\leq \|A\|_{F}^{2} \ ,
\end{equation}
and the left inequality becomes an equality if and only if~:
\begin{equation}
\label{570-3}
  \spann Z = \spann \{ \ww_{1}\dots \ww_{\nl}\} \ ,
\end{equation}
where $\ww_{1}, \dots \ww_{\nl}$ are the $\nl$ first right singular vectors of $A$.
\end{Theorem}

\subsection{Proof of Proposition \ref{pro resolution cl sgb}}
\label{proof of pro resolution cl sgb}

For any $X=[x_{1} \dots x_{\nl}]\in \stns$ and any $i=1\dots\nvg \, , \, j=1\dots\nl$ we introduce the polar decomposition (cf (\ref{533})) of the vector $A_{i}^{T}x_{j}\in \R^{\dv_{i}}$~:
\begin{equation}
\label{400-10}
  a_{i}^{T}x_{j} = u_{ij} \,\alpha_{ij} \ , \ \mbox{with} \quad \|u_{ij}\|=1 \quad , \quad 
  \alpha_{ij}\geq 0 \ ,
\end{equation}
and define for $j=1 \dots \nl$ the vectors $t_{j} =(t_{ij} \,,\, i=1 \dots \nvg)$ of $\R^{\nvs}$ by~:
\begin{equation}
\label{402-10}
\left\{ \begin{array}{ lcl}
  t_{ij} &=&  u_{ij} [\alpha_{ij}-\gamma_{j}]_{+} 
  \in \R^{\dv_{i}} \quad, \quad i= 1 \dots \nvg   \ . \\
 \displaystyle  \|t_{j}\|^{2} & = & \displaystyle \sum_{i = 1}^{\nvg}  [\alpha_{ij}-\gamma_{j}]_{+}^{2} \ .
\end{array} \right.
\end{equation}

We give first an analytical solution to the inner maximization problem in (\ref{350})~: we show that
\begin{equation}
\label{416}
 \forall X \in \stns \quad , \quad  
 \max_{Z \in \bpnvs}  \sum_{j=1\dots\nl}\mu_{j}^{2}
   \big[ x_{j}^{T}Az_{j} -\gamma_{j}\|z_{j}\|_{1} \big]_{+}^{2} 
  = F(X) \mbox{ given by }
 (\ref{404}) \ ,
\end{equation}
which proves the first part of Proposition \ref{pro resolution cl sgb}. 

So let \emph{$X$ be a given point on the Stiefel manifold $\stns$}. Then~:
\begin{eqnarray}
\hspace{-1.5em}
\max_{Z \in \bpnvs}  \sum_{j=1\dots\nl}\mu_{j}^{2}
   \big[ x_{j}^{T}Az_{j} -\gamma_{j}\|z_{j}\|_{1} \big]_{+}^{2} & = & \max_{\|z_{j}\|\leq 1\ ,\ j=1\dots\nl}  
				\sum_{j=1}^{\nl}\mu_{j}^{2}  
				\big[ x_{j}^{T}Az_{j} - \gamma_{j}\|z_{j}\|_{1}\big]_{+}^{2}    
			\label{420} 	\\
 & = &  	\sum_{j=1}^{\nl}\mu_{j}^{2} 
 			\max_{\|z_{j}\|\leq 1} 
				\big[ x_{j}^{T}Az_{j} - \gamma_{j}\|z_{j}\|_{1}\big]_{+}^{2} \ ,    
				\label{421}  
\end{eqnarray}
But $t\leadsto [t]_{+}^{2}$ is a monotonously increasing function, hence~:
\begin{eqnarray}
 \max_{Z \in \bpnvs}  \sum_{j=1\dots\nl}\mu_{j}^{2}
   \big[ x_{j}^{T}Az_{j} -\gamma_{j}\|z_{j}\|_{1} \big]_{+}^{2} & = &  	\sum_{j=1}^{\nl}\mu_{j}^{2} 
 			\big[  \max_{\|z_{j}\| \leq 1} (x_{j}^{T}Az_{j} 
			- \gamma_{j}\|z_{j}\|_{1})\big]_{+}^{2} \ .    \label{423} 
\end{eqnarray}
The max in the right-hand side of (\ref{423}) is certainly positive, as
$z_{j}=0$ belongs to the admissible set $\{z \, | \,\|z\| \leq 1\}$, and (\ref{423}) becomes~:
\begin{eqnarray}
  \max_{Z \in \bpnvs}  \sum_{j=1\dots\nl}\mu_{j}^{2}
   \big[ x_{j}^{T}Az_{j} -\gamma_{j}\|z_{j}\|_{1} \big]_{+}^{2} & = &  	\sum_{j=1}^{\nl}\mu_{j}^{2}  \Big (
 			 \max_{\|z_{j}\| \leq1} (x_{j}^{T}Az_{j} 
			- \gamma_{j}\|z_{j}\|_{1}) \Big) ^{2} \ ,    \label{424} 
\end{eqnarray}
Hence the inner maximization problem (\ref{420}) reduces to the solution of $\nl$ independant optimization problems with respect to $z_{j},j=1\dots\nl$ . So we drop the index $j$, and consider, \emph{given $x \in \R^{\ns}$ with $\|x\|=1$}, the  optimization problem~:
\begin{eqnarray}
\label{430}
  z^{*} &=& \arg \max_{\|z\| \leq 1} ( x^{T}Az - \gamma \|z\|_{1} )   \\
  \label{431}
  	&=& \arg \max_{\|z_{1}\|^{2}+\dots+\|z_{p}\|^{2}  \leq 1} 
	\sum_{i=1}^{\nvg} \big( x^{T}a_{i}z_{i} - \gamma \|z_{i}\| \big) \ , 
\end{eqnarray}
where  the $z_{i}\in \R^{\dv_{i}}$ are the loadings associated to each group variable.
We introduce the polar decomposition  (cf (\ref{533})) of $z_{i}$ in $\R^{p_{i}}$~:
\begin{equation}
 \label{437-1}
   z_{i} = v_{i} \,\beta_{i} \ , \ \mbox{with} \quad \|v_{i}\|=1 \quad , \quad 
  \beta_{i}\geq 0 \ .
\end{equation}
and replace the search for $z^{*}$ by that for $v_{i}^{*},\beta_{i}^{*}, i=1 \dots \nvg$.
Then equation  (\ref{431}) becomes~:
\begin{equation}
\label{437-4}
\hspace{-0.1em}
  (v_{i}^{*},\beta_{i}^{*} , i=1 \dots \nvg) =\arg \max_{
  \|v_{i}\|=1, i=1 \dots \nvg
 } \!\!
\max_{
\begin{array}{c}
    \sum_{i=1 \dots \nvg} \beta_{i}^{2}  \leq 1    \\
      \beta_{i} \geq 0 \ , \ i=1 \dots \nvg
\end{array}
 }
\!\!	\sum_{i=1}^{\nvg} \max_{\|v_{i}\|=1} \big(  \alpha_{i}\beta_{i} u_{i}^{T}v_{i} - \gamma \beta_{i} \big) \ . \hspace{-2em}
\end{equation}	
The first maximum is obviously achieved for~:
\begin{equation}
\label{437-5}
  v_{i}^{*}=u_{i} \quad , \quad i=1 \dots \nvg \ ,
\end{equation}
and (\ref{437-4}) reduces to~:
\begin{equation}
\label{437-6}
   (\beta_{i}^{*} , i=1 \dots \nvg)     =\arg \max_{
   \begin{array}{c}
     \sum_{i=1 \dots \nvg} \beta_{i}^{2}  \leq 1     \\
      \beta_{i} \geq 0 \ , \ i=1 \dots \nvg
\end{array}
   } 
	\sum_{i=1}^{\nvg}  ( \alpha_{i} - \gamma ) \beta_{i} \ ,
\end{equation}
Define~:
\begin{equation}
\label{440}
     I_{+}= \{i = 1\dots\nvg\ | \ \alpha_{i}-\gamma > 0 \}  \ .
  \end{equation}
\begin{itemize}
  \item either~: $I_{+}= \emptyset$, and~:
\begin{equation}
\label{446}
  \beta^{*}=0
\end{equation}
is a trivial solution of (\ref{437-6}) - but non-necessarily unique if $\alpha_{i}-\gamma=0$ for some $i$. 
  \item or~: $I_{+} \neq \emptyset$.
We check first that in this case~:
\begin{equation}
\label{445} 
  \beta_{i}^{*} = 0 \quad \forall i \notin I_{+}  \ .
\end{equation}
For that purpose, suppose that $\beta_{\ell}^{*} > 0$ for some $\ell \notin I_{+}$, and let $k$ be an index of $I_{+}$.
One can define $\tilde{\beta}^{*}$ by  $\tilde{\beta}_{i}^{*}=\beta_{i}^{*}$ for $i\neq k , \ell$, $\tilde{\beta}_{\ell}^{*}= 0$, and $\tilde{\beta}_{k}^{*} > \beta_{k}^{*} $ such that
$ \| \tilde{\beta}^{*} \|=\| \beta^{*}\| \leq 1 $. Then~:
\begin{eqnarray}
( \alpha_{\ell} - \gamma ) \beta_{\ell}^{*}& \leq & 0 =
( \alpha_{\ell} - \gamma ) \tilde{\beta}_{\ell}^{*}  \ ,   \nonumber \\
( \alpha_{k} - \gamma ) \beta_{k}^{*}& < & 
( \alpha_{k} - \gamma ) \tilde{\beta}_{k}^{*}  \ , \nonumber
\end{eqnarray}
which contradicts the fact that $\beta^{*}$ is a maximizer, and ends the proof of (\ref{445}).

We can now restrict the search to the 
  $(\beta_{i}^{*},i \in I_{+})$, so (\ref{437-6}) simplifies to~:
\begin{eqnarray}
  (\beta_{i}^{*} , i\in I_{+})   &=&
    \arg \max_{
   \begin{array}{c}
     \sum_{i\in I_{+}} \beta_{i}^{2}  \leq 1     \\
      \beta_{i} \geq 0 \ , \ i\in I_{+}
\end{array}
   } 
	\sum_{i\in I_{+}} ( \alpha_{i} - \gamma ) \beta_{i} \ , \label{455} \\
 &=&  \arg \max_{
   \begin{array}{c}
     \sum_{i\in I_{+}} \beta_{i}^{2}  \leq 1  
\end{array}
   } 
	\sum_{i\in I_{+}} ( \alpha_{i} - \gamma ) \beta_{i} \ , 
    \label{456}
\end{eqnarray}
where the last equality holds because the coefficients $\alpha_{i}-\gamma$ of $\beta_{i}$ are positive for $i \in I_{+}$. Hence the solution $\beta^{*}$ of (\ref{437-6}) is given, when $I_{+}\neq \emptyset$,  by~:
\begin{equation}
 \label{460}
        \beta_{i}^{*} =  \frac{\big[ \, \alpha_{i}-\gamma \big]_{+}}
     { \big( \sum_{i=1\dots \nvg} \big[ \, \alpha_{i}-\gamma \big]_{+}^{2}\big)^{1/2}} 
     \quad,\quad i= 1 \dots \nvg \ ,
 \end{equation}
 \end{itemize}
 Returning to the $z$ unknowns one obtains, using (\ref{437-1})(\ref{437-5})(\ref{446})(\ref{460})~:
\begin{equation}
\label{461}
z^{*} = \left\{\begin{array}{ll}
   0   &  \mbox{ if } I_{+}=\emptyset \ , \\
   z_{i}^{*}= u_{i} \beta_{i}^{*} \, ;\, i=1 \dots,\nvg &    \mbox{ if } I_{+} \neq \emptyset \ ,
\end{array} \right.
\end{equation}
 and in both cases the maximum of the optimization problem (\ref{430}) is given by~:
 \begin{equation}
 \label{462}
 \max_{\|z\| \leq 1} ( x^{T}Az - \gamma \|z\|_{1} )  \hspace{1,5em}  
 = \hspace{1,5em} \big( \sum_{i=1}^{\nvg} [ \alpha_{i} - \gamma]_{+}^{2} 
  \big)^{1/2} \ .
\end{equation}
Reintroducing the $j$ indices, the solution of the inner maximization problem (\ref{420}), for a given  $X \in \stns$, is, using its reformulation (\ref{424}) together with (\ref{461}), (\ref{462}) and the notation $t_{j}^{*} \in \R^{\nvs} $ defined in (\ref{404})~:
\begin{equation}
\label{468}
 \forall  j= 1 \dots \nl \quad , \quad  z_{j}^{*}= \left\{\begin{array}{ll}
   0   & \mbox{ if } t_{j}^{*}=0   \ ,\\
    t_{j}^{*}/\|t_{j}^{*}\|   &   \mbox{ if } t_{j}^{*} \neq 0 \ ,
\end{array}\right.
\end{equation}
\begin{equation}
\label{464}
\max_{Z \in \bpnvs}  \sum_{j=1\dots\nl}\mu_{j}^{2}
   \big[ x_{j}^{T}Az_{j} -\gamma_{j}\|z_{j}\|_{1} \big]_{+}^{2}
=  \sum_{j=1\dots\nl} \mu_{j}^{2}  \|t_{j}^{*}\|^{2}
 = F(X)\ .
\end{equation}
The last equation proves (\ref{416}) , and hence part 1 of the theorem. Then (\ref{468}) gives (\ref{412}) when $X$ is a solution $X^{*}$ of (\ref{404}), and part 2 is proved.

We prove now point 3 of the proposition~: let the sparsity parameters $\gamma_{j}$ satisfy {\eqref{414}. Hence there exists $\ell \in 1 \dots \nl$ and $k \in 1 \dots \nvg$ such that~:
\begin{equation}
\label{465}
  \gamma_{\ell} < \|a_{k} \|_{2} =  \|a_{k}^{t} \|_{2}  \ .
\end{equation}
By definition of the matrix norm $\|.\|_{2}$, there exists $X \in \stns$ such that $x_{\ell}$ satisfies:~:
\begin{equation}
\label{466}
  \gamma_{\ell} < \|a^{t}_{k} \, x_{\ell} \| = \alpha_{k \ell} \ .
\end{equation}
Then (\ref{402}) gives~:
\begin{equation}
\label{467}
 \|t_{\ell}\|^{2} \geq (\alpha_{k\ell}-\gamma_{\ell})^{2} >0  \quad \Longrightarrow \quad 
 t_{\ell}\neq 0 \ ,
\end{equation}
and point 3 is proven. \cqfd

\subsection{Proof of Lemma \ref{lem 1}}
\label{proof of lem 2}

The left equality in \eqref{546} follows from $\pr_{Z}= Z(Z^{T}Z)^{-1}Z^{T}$, which implies that  
\newline \hspace*{3em}$\varsubspace(AZ) \egaldef \|A\pr_{Z}\|_{F}^{2} = \|AZ(Z^{T}Z)^{-1/2}\|_{F}^{2}
= \tr\{Z^{T}A^{T}A Z(Z^{T}Z)^{-1}\}$.

\noindent The  inequality in \eqref{546} (right) as well as  (\ref{546-2}) follow immediately from the properties of the generalized Rayleigh quotient $\tr\{ Z^{T}A^{T}A Z(Z^{T}Z)^{-1})\}$ recalled in Theorem \ref{thm GRQ} of the Appendix. 
Property 1 follows from the definition of $\varsubspace$, and condition \eqref{510a} of property 2 follows from (\ref{546}). It remains to prove \eqref{546-1} and  \eqref{546-4} which shows that property 3 does not hold.

Let $Y=AZ$ be orthogonal components~:
 \begin{equation}
\label{900}
  \langle y_{j},y_{k}\rangle = 0 \  , \ j,k=1 \dots \nl,j\neq k
\end{equation}
corresponding to unit norm loadings~:
\begin{equation}
\label{901}
 \|z_{j}\| =1 \ j=1 \dots \nl \ ,
\end{equation}
and define $X,T$ by~:
\begin{equation}
\label{904}
  x_{j}= y_{j}/\|y_{j}\| \quad , \quad t_{j}=z_{j}/\|y_{j}\| 
  \quad , \quad j=1 \dots \nl \ ,
\end{equation}
so that~:
\begin{equation}
\label{908}
  X^{T}X = I_{\nl} \ .
\end{equation} 
 Then on one side one has~:
 \begin{equation}
\label{912}
  \|Y\|_{F}^{2} =  \sum_{j=1 \dots \nl} \|y_{j}\|^{2} 
  = \sum_{j=1 \dots \nl}1/\|t_{j}\|^{2}
  = \tr\{\diag^{-1}(T^{T}T)  \} \ ,
\end{equation}
 and on the other side, as $Y$ and $X$ span the same subspace~:
\begin{equation}
\label{914}
  \varsubspace Y = \varsubspace X 
  =  \tr\{ (X^{T}X)(T^{T}T)^{-1} \} 
  =  \tr\{(T^{T}T)^{-1} \} 
\end{equation}
 Formula \eqref{546-1} will be proved if we show that~:
 \begin{equation}
\label{916}
  \tr\{\diag^{-1}(T^{T}T)  \}  \leq  \tr\{(T^{T}T)^{-1} \} \ .
\end{equation}
We use for that an idea taken from \cite{M1969}, and perform
 a QR-decomposition of $T$.
By construction, the diagonal elements of $R$ satisfy~:
\begin{equation}
\label{918}
   0 < r_{i,i} \leq \|t_{i}\| \ .
\end{equation}
Then~:
\begin{eqnarray}
\label{920}
  T^{T}T & = & R^{T}   Q^{T} Q\, R 
            =  R^{T}  R  \ ,\\
    (T^{T}T)^{-1} & = & R^{-1} (R^{T})^{-1}  =
    R^{-1} (R^{-1})^{T} \ ,         
\end{eqnarray}
where $ R^{-1}$ satisfies~:
\begin{equation}
\label{922}
   R^{-1} = \mbox{upper triangular matrix} \quad , \quad
   [R^{-1}]_{i,i} = 1/ r_{i,i} \ .
\end{equation}
Hence  the diagonal element of $(T^{T}T)^{-1}$ are given by~:~:
\begin{eqnarray}
\label{924}
  \big[(T^{T}T)^{-1}\big]_{i,i} & = & \big[ R^{-1} (R^{-1})^{T} \big]_{i,i}  \\
  \nonumber
      & = &  [R^{-1}]_{i,i}^{2} + \sum_{j>i} [R^{-1}]_{i,j}^{2} \\
      \nonumber
      & \geq &  [R^{-1}]_{i,i}^{2}  = 1/r_{i,i}^{2}  \geq 1/\|t_{i}\|^{2} \ .
\end{eqnarray}
which gives (\ref{916}) by summation over $i=1 \dots \nl$, and \eqref{546-1} is proved.

We prove now the left-to-right implication in \eqref{546-4}~: let the orthogonal components $y_{j},j=1\dots \nl$ satisfy $ \|Y\|_{F}^{2} = \varsubspace Y$.
Then \eqref {912} \eqref{914} imply that all inequality in \eqref{924} are equalities~:
\begin{enumerate}
  \item first inequality~: $ [R^{-1}]_{i,j}^{2}=0 \text{ for all } j>i \quad \Rightarrow  \quad
   \text{ $R^{-1}$ and hence $R$ are diagonal } $
  \item second inequality~: $1/r_{i,i}^{2}  = 1/\|t_{i}\|^{2} \quad \Rightarrow \quad
  \text{$R$ is diagonal}$
\end{enumerate} 
But $R$ diagonal implies that the $t_{j}$ - and hence also the loadings $z_{j}$ - are orthogonal, which together with the hypothesis of orthogonal components $y_{j}$, implies that $(y_{j}/\|y_{j}\|,z_{j})$ are pairs of singular vectors of $A$, and ends the proof of the left-to-right implication.

Conversely, let $z_{j}=v_{\ell(j)},j=1\dots\nl$. Then 
$\|Y\|^{2} = \sum_{,j=1\dots\nl} \sigma_{\ell(j)}^{2}=\varsubspace Y$, where the last equality follows from the Generalized Rayleigh Quotient formula  \eqref{546} for the subspace variance.
This ends the proof of the lemma \cqfd

\subsection{Proof of Lemma \ref{lem 3}}  
\label{proof of lem 3}

Let $\E = A\, \snvs$ be the $\ns$-dimensional ellipsoid  image by $A$ of the unit sphere $\snvs \subset \R^{\nvs}$, and~:
\begin{equation}
\label{800}
  \EX = \E \cap  \spann Y = \E \cap  \spann X  
\end{equation}
the $\nl$-dimensional ellipsoid, trace of $\E$ on the subspace spanned both by the given components $Y$ and the chosen basis $X$. By construction one has~:
\begin{equation}
\label{802}
  y_{j} \in \EX \quad , \quad j=1 \dots \nl \ ,
\end{equation}
and the modified components $Y\p$ defined by projection satisfy, c.f. (\ref{532})~:
 \begin{equation}
\label{804}
  \|y\p_{j}\| = | \langle y_{j},x_{j}\rangle | \leq \ve_{j} \egaldef 
  \max_{\displaystyle y \in \EX} \langle y , x_{j}\rangle  \ , \  j=1\dots \nl \ ,
\end{equation}
so that~:
\begin{equation}
\label{805}
  \|Y\p\|_{F}^{2} \leq \ve_{1}^{2}+ \dots+ \ve_{\nl}^{2} \ .
\end{equation}
We can now ``box'' the ellipsoid $\EX$ in the parallelotope $\PX$ of $\spann X$ defined by~:
\begin{equation}
\label{806}
  \PX = \big \{  y\in \spann X \   | \  -\ve_{j} \leq \langle y,x_{j}\rangle \leq + \ve_{j}  \ , \  j=1\dots \nl \big \} \ ,
\end{equation}
(see figure \ref{fig 2}).
By construction, one can draw from each of the $2^{\nl}$ vertices of $\PX$ $\nl$ orthogonal hyperplanes tangent to the ellipsoid $\EX$, which implies that they are all on the orthoptic or Cartan sphere of the ellipsoid, whose radius is known to be the sum of the squares of the semi-principal axes $\sigma^{X}_{j}, j = 1 \dots \nl$ of $\EX$ (see for example the textbook \cite{PT2005}).

\begin{figure}[h]
\begin{center}
\centerline{\resizebox{28em}{!}{\includegraphics{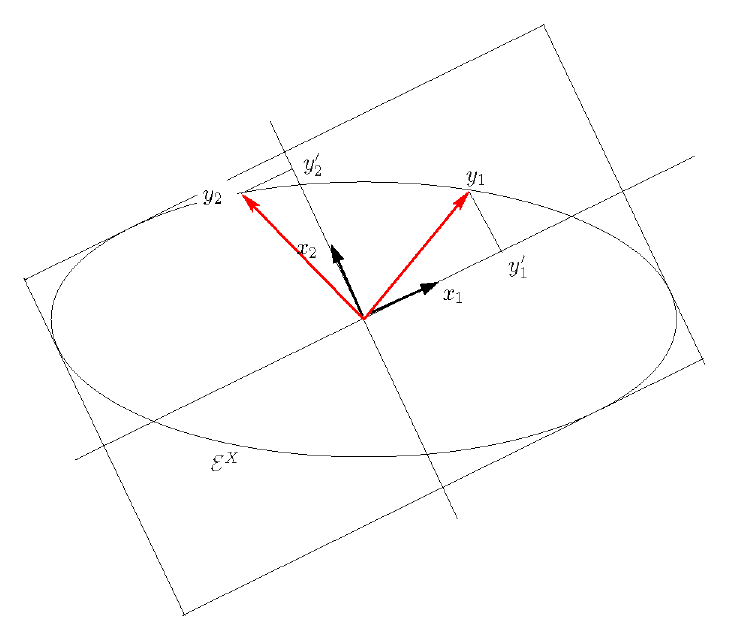}}}
\caption{Illustration of the upper bound to $\|Y\p\|_{F}^{2}$ in $\spann Y$ when $Y\p$ is defined by projection.}
\label{fig 2}
\end{center}
\end{figure}

Hence~:
\begin{equation}
\label{808}
  \ve_{1}^{2}+ \dots + \ve_{\nl}^{2} = (\sigma^{X}_{1})^{2} + \dots +(\sigma^{X}_{\nl})^{2} \ .
\end{equation}
Let then $y^{X}_{1}\dots y^{X}_{\nl}$ be vectors whose extremity are points of $\EX$ located on its principal axes, so that~:
\begin{equation}
\label{810}
 \|y^{X}_{j}\|=\sigma^{X}_{j}\ ,\ j=1\dots \nl \quad , \quad 
 \langle y^{X}_{i},y^{X}_{j}\rangle = 0\ , \ i,j=1\dots \nl , i\neq j \ .
\end{equation}
Property \eqref{546-1}  of Lemma \ref{lem 1} applied to $Y=Y^{X}$ gives:~:
\begin{equation}
\label{812}
 (\sigma^{X}_{1})^{2} + \dots + (\sigma^{X}_{\nl})^{2} =  \|Y^{X}\|^{2} \leq
  \varsubspace \, Y^{X} \leq \varpca \ .
\end{equation}
Combining inequalities (\ref{805}) (\ref{808}) (\ref{812}) proves the inequality (\ref{534-2}). 

We prove now the left-to-right implication in  \eqref{534-6}. Suppose that, for 
some $X$, the equality $\varproj Y =\varpca$ holds. Then necessarily~:
\begin{itemize}
  \item    equality holds in \eqref{812},
which requires that the loadings $Z$ span the  subspace $V_{\nl}$ of the $\nl$ first right singular vectors, which proves the right part of \eqref{534-6}, top.
  \item and that equality holds in \eqref{805}, which implies that for $j\neq k$ the normals to $\EX$ at $y_{j}$ and $y_{k}$ are orthogonal (see Figure \ref{fig 2}). 
  The restriction $A_{\nl}$ of $A$ to $\spann V_{\nl}$ is an isomorphism from $\spann V_{\nl}$ to $\spann \, U_{\nl}$, hence~:
\begin{equation}
\label{814}
\EX = \{ y \in \spann \, U_{\nl} \  | \  \|A_{\nl}^{-1}y\|^{2} = 1 \} \ .
\end{equation}
 A normal $n(y)$ to $\EX$ at $y$ is then~:
\begin{equation}
\label{816}
 n(y)= \nabla_{y} \big(  \|A_{\nl}^{-1}y\|^{2} - 1 \big)
 = 2(A_{\nl}^{-1})^{T}A_{\nl}^{-1}y =  2(A_{\nl}^{-1})^{T} z
 = 2 \diag\{ \frac{1}{\sigma_{1}} \dots \frac{1}{\sigma_{\nl}} \} z \   ,
\end{equation}
where vectors and matrices are written on the singular bases $U_{\nl}$ and $V_{\nl}$.
The orthogonality of $n(y_{j})$ and $n(y_{k})$ proves the right 
part of \eqref{534-6}, bottom.
\end{itemize} 
Conversely, let the right part (top) of \eqref{534-6} hold. This implies that the half axes of $\EX$ are $\sigma_{1}\dots\sigma_{\nl}$. Then the right part, bottom, implies that the normal $n_{j}$ to $\EX$ at $y_{j},j=1\dots\nl$ are orthogonal. 
So one can box $\EX$ with a parallelotope $\PX$ with axes parallel to the normals $n_{j}$, and define $X$ as the orthonormal basis along its axes. Then the same reasonning as above for the proof of (\ref{534-2}) shows that 
$\varproj Y = \varpca$, which ends the proof of the lemma.
\cqfd

\bibliographystyle{plain}
\bibliography{BlockSparseGroupACP}

\end{document}